\documentclass[10pt,journal,compsoc,lettersize]{IEEEtran}
\usepackage{amsmath,amsfonts}
\usepackage{array}
\usepackage[caption=false,font=normalsize,labelfont=sf,textfont=sf]{subfig}
\usepackage{textcomp}
\usepackage{stfloats}
\usepackage{url}
\usepackage{verbatim}
\usepackage{graphicx}
\def\BibTeX{{\rm B\kern-.05em{\sc i\kern-.025em b}\kern-.08em
    T\kern-.1667em\lower.7ex\hbox{E}\kern-.125emX}}
\usepackage{balance}

\usepackage{multirow}
\usepackage{colortbl}
\usepackage{wrapfig}
\usepackage{enumitem}
\usepackage{array}
\usepackage{float}
\usepackage{xcolor}
\usepackage{bbding}
\usepackage{makecell}
\usepackage{amsmath}
\usepackage{booktabs}
\usepackage{amssymb}
\usepackage{pifont}

\usepackage{algpseudocode}
\usepackage[table]{xcolor}

\newcommand{\cmark}{\cellcolor{green!35}\ding{51}}
\newcommand{\xmark}{\cellcolor{red!25}\ding{55}}
\newcolumntype{L}[1]{%
    >{\raggedright\arraybackslash}p{#1}%
}

\definecolor{oursbg}{RGB}{240,244,248}
\definecolor{groupbg}{RGB}{247,245,241}

\newcommand{\best}[1]{\textbf{#1}}
\newcommand{\second}[1]{\underline{#1}}
\newcommand{\bad}[1]{\textcolor{red!70!black}{#1}}

\newcommand{\hrs}[1]{#1\,{\textbf{\scriptsize h}}}
\newcommand{\mins}[1]{#1\,{\textbf{\scriptsize min}}}
\newcommand{\secs}[1]{#1\,{\textbf{\scriptsize s}}}

\newcounter{CeFlunealgorithm}
\renewcommand{\theCeFlunealgorithm}{\arabic{CeFlunealgorithm}}
\algrenewcommand\alglinenumber[1]{\scriptsize\color{black!55}#1:}

\usepackage{hyperref}
\hypersetup{colorlinks=true, citecolor=blue}

\begin{document}
\bstctlcite{BSTcontrol}
\title{Robust Concept Erasure in Text-to-Image Models via Simulating Brain Functional Localization}
\author{Manyi Li, Yufan Liu, Bing Li, Xia Wu, Ziyu Li, Weiming Hu
\thanks{\textit{Manyi Li and Yufan Liu are co-first authors.
Corresponding author: Bing~Li.}}
\IEEEcompsocitemizethanks{
  \IEEEcompsocthanksitem  M. Li, Y. Liu, B. Li, and W. Hu are with the Beijing Key Laboratory of Super Intelligent Security of Multi-Modal Information, State Key Laboratory of Multimodal Artificial Intelligence Systems, Institute of Automation, Chinese Academy of Sciences, Beijing 100190, China (e-mail: limanyi25@mails.ucas.ac.cn; yufan.liu@ia.ac.cn; bli@nlpr.ia.ac.cn; wmhu@nlpr.ia.ac.cn).
  \IEEEcompsocthanksitem Z. Li and X. Wu are with the School of Computer Science and Technology, Beijing Institute of Technology, Beijing 100190, China (e-mail: ziyuli@bit.edu.cn; wuxia@bit.edu.cn).
  \IEEEcompsocthanksitem M. Li is also with the School of Advanced Interdisciplinary Sciences, University of Chinese Academy of Sciences, Beijing 101408, China.
}%
}

\markboth{IEEE TRANSACTIONS ON PATTERN ANALYSIS AND MACHINE INTELLIGENCE}{}
\maketitle

\begin{abstract}
Concept erasure in text-to-image models aims to prevent a target concept from appearing in generated images while preserving general image generation capability. However, erased concepts can still be reactivated by attacks. Prior attacks mainly search text or image input spaces, leaving the initial latent space underexplored. To address this gap, we introduce IVO (Initial latent Variable Optimization), a new concept reactivation attack that changes the starting point of image generation by optimizing the initial latent. Comprehensive experiments across 8 erasure methods show that IVO reactivates erased concepts more effectively than the attack baselines, revealing a blind spot in erasure robustness evaluation. This motivates us to localize model components associated with the target concept and then treat them as intervention sites for more robust concept erasure. We propose CeFlune (Concept Erasure via Functional Localization of Unit Neurons), which adapts brain functional localization to text-to-image models. CeFlune defines linear modules as Unit Neurons and localizes concept-responsive ones by comparing their response trajectories in the model’s task and resting states. It then updates the localized Unit Neurons to constrain their responses and the erased model’s predictions. Across U-Net diffusion and flow-matching transformer architectures, CeFlune achieves a favorable balance between concept erasure and general image generation capability in comparisons with up to 15 baseline methods. Under IVO and 4 other evaluated attacks, CeFlune also shows lower concept reactivation.
\end{abstract}


\section{Introduction} \label{sec:Introduction}
Text-to-image models can generate realistic and diverse images from text prompts. However, practical applications may require restricting sexually explicit imagery, copyrighted content, or unauthorized portraits. Concept erasure addresses these needs through post-training intervention to erase concepts in a pretrained model \cite{ESD}. The term ``concept'' refers to a content category, identity, or visual attribute that can be described in text and recognized in generated images.

Concept erasure is commonly framed around two goals: erasing a concept and preserving general image generation capability. Existing methods implement concept erasure through parameter fine-tuning \cite{ESD}, closed-form weight updates \cite{UCE}, saliency-guided parameter selection \cite{SalUn}, and learned gates that control the scope of erasure \cite{CPE}. However, successful erasure with standard inputs does not establish robustness to concept reactivation attacks, which search for inputs that make an erased model generate images containing the erased concept.

\begin{figure}[!t]
    \centering
    \includegraphics[width=\columnwidth]{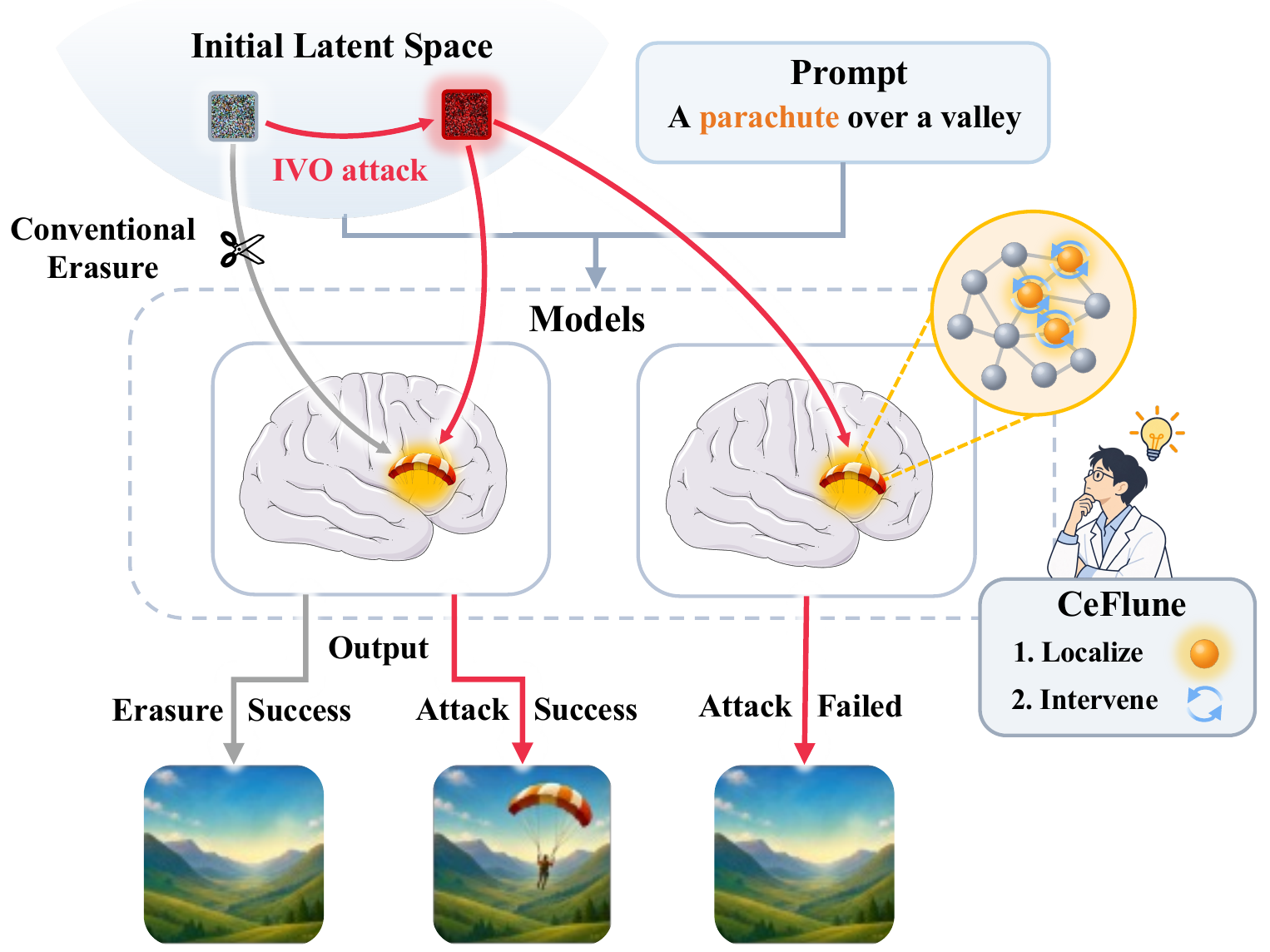}
    \caption{Concept reactivation by IVO and its mitigation by CeFlune. With the prompt fixed, conventional erasure appears to erase the parachute concept under standard generation, but an IVO-optimized initial latent reactivates it. CeFlune localizes and intervenes on concept-responsive Unit Neurons, which prevents the illustrated IVO attack from reactivating the parachute concept.}
    \label{fig:teaser}
\end{figure}

Previous concept reactivation attacks mainly optimize text prompts \cite{P4D}, text embeddings \cite{CI}, or image inputs \cite{RECALL} used to guide generation, but overlook another essential input to the generation process: the initial latent. As the generation starting state, the initial latent can be directly manipulated in a high-dimensional, continuous latent space, extending the adversarial search beyond the text and image input spaces. To address this gap, we introduce \textbf{IVO} (Initial latent Variable Optimization), a two-stage concept reactivation attack. IVO first inverts a reference image containing the erased concept to obtain an initial latent. It then optimizes this latent using a concept-reactivation loss and a semantic-anchoring loss. The concept-reactivation loss encourages the erased model’s predictions to match those of the original model for the same target prompt, while the semantic-anchoring loss constrains semantic drift to support the generated image consistent with the target prompt.

Fig.~\ref{fig:teaser} shows that IVO can reactivate a concept after conventional erasure by changing the initial latent. This prompts us to locate the model components associated with forming and transforming representations of the target concept, a process known as concept localization. These components can then serve as intervention sites for an erasure method designed to resist concept reactivation attacks. Brain functional localization pursues a goal similar to concept localization: it seeks to identify brain regions associated with a particular task. Task-based functional magnetic resonance imaging (fMRI) is a widely used technique for this purpose \cite{park2020mapping}. It records brain signals while a participant performs the task (task state) and while they rest (resting state). Comparing these signals reveals task-evoked responses that help identify task-related brain regions.

We propose \textbf{CeFlune} (Concept Erasure via Functional Localization of Unit Neurons), a concept erasure method that simulates brain functional localization in text-to-image models. CeFlune defines each linear module as a Unit Neuron and its response as a signal of the module’s activity. A prompt describing the concept induces the model’s task state, while empty text induces its resting state. CeFlune first measures Unit Neuron responses across generation steps, forming response trajectories for the task and resting states. By comparing these trajectories, it selects concept-responsive Unit Neurons and then updates their parameters to perform localized concept erasure. Specifically, CeFlune jointly optimizes a prediction-space loss that guides concept erasure while helping preserve general image generation capability and a task-to-rest response alignment loss that anchors the selected Unit Neurons’ task-state responses to the frozen original model’s resting-state responses.

We evaluate 8 existing erasure methods with IVO across diverse concept categories and find that it effectively reactivates erased concepts. These results reveal vulnerabilities of the evaluated erasure methods in the initial latent space, demonstrating the value of IVO for evaluating the robustness of concept erasure. Across 6 concepts, CeFlune achieves strong erasure while preserving general image generation capability when compared with 15 erasure baselines on SDv1.4 and 11 on SDv3. It also achieves the lowest average attack success rate among the erasure baselines under 5 concept reactivation attacks. Results on other model architectures further support the applicability of CeFlune. Our main contributions are:

\begin{itemize}[leftmargin=10pt]
\item We introduce IVO, which reactivates erased concepts by adversarially optimizing the initial latent. IVO extends robustness evaluation of concept erasure to the latent space, exposing vulnerabilities in existing erasure methods.
\item We propose CeFlune, which simulates brain functional localization to locate concept-responsive Unit Neurons. It updates their parameters to erase the concept.
\item Comprehensive experiments show that IVO effectively reactivates concepts erased by existing erasure methods. Across different model architectures, CeFlune achieves a balance between concept erasure and preservation of general image generation capability, while showing stronger overall resistance to IVO and 4 other evaluated attacks.
\end{itemize}

 \section{Related Work} \label{sec:Related Work}
We review three connected lines of research: methods that erase concepts from text-to-image models, attacks that attempt to reactivate erased concepts, and studies that localize concepts in text-to-image models.

\subsection{Concept Erasure Methods} \label{sec:Concept Erasure Methods}
Concept erasure can be viewed as a task-specific form of machine unlearning for text-to-image generation. It aims to prevent a concept from appearing in generated images by modifying a pretrained text-to-image model.

Early concept erasure methods mainly change pretrained weights. ESD \cite{ESD} fine-tunes model parameters so that the model itself steers generation away from the target concept, while AC \cite{AC} maps the target concept toward a broader or predefined replacement concept. FMN \cite{FMN} uses cross-attention scores, which describe the relation between image features and text condition, as a fine-tuning objective. Other methods avoid iterative training. UCE \cite{UCE} computes a closed-form update to the model's attention projections, and MACE \cite{MACE} combines such closed-form refinement with low-rank adaptation (LoRA) fine-tuning to erase many concepts at a time. SalUn \cite{SalUn} first identifies salient weights and only updates them.

A second group of methods emphasizes preserving general image generation capability, restricting erasure to concept-relevant image regions, or scaling concept erasure across multiple concepts. SPM \cite{SPM} learns transferable one-dimensional adapters, while Receler \cite{Receler} inserts adapters after cross-attention layers and trains them with concept-localized regularization and adversarial prompt learning. CPE \cite{CPE} uses residual attention gates to focus erasure on the specified concept while preserving other concepts. GLoCE \cite{GLoCE} uses a training-free gated low-rank module to erase a concept only at the image region where it appears, preserving co-occurring content elsewhere. NLCE \cite{NLCE} combines spatially localized erasure with representation-space modulation to better protect semantically related concepts. These forms of localization concern where concept erasure takes effect in generated images, rather than which model components should be selected for intervention. More recent work improves the scalability of concept erasure through closed-form parameter updates, as in SPEED \cite{SPEED} and ScaPre \cite{ScaPre}. 

These methods differ in the cost of erasure and the scope of parameter intervention, but most of them share a common practice: choosing an intervention site in advance, rather than first asking which parts of the model are associated with the concept. CeFlune addresses this question by defining linear modules as Unit Neurons, identifying those that contribute to forming representation of the concept, and fine-tuning the selected Unit Neurons. Because linear modules are a basis for both U-Net and Transformer backbones, CeFlune can be applied across different generative architectures. Table~\ref{tab:model_compatibility} shows that CeFlune is the only method covering all five representative model families across diffusion and flow matching.

\subsection{Concept Reactivation Attacks} \label{sec:Concept Reactivation Attacks}
Concept reactivation attacks search for inputs that reactivate a previously erased concept. Existing approaches differ in the input space they optimize and the type of access they require to the erased model. 

CI \cite{CI} learns new text embeddings that retrieve the erased concept without changing the erased model's weights. P4D \cite{P4D}, Ring-A-Bell \cite{Ring-A-Bell}, and UnlearnDiffAtk \cite{UnlearnDiffAtk} use different optimization and red-teaming procedures to search for prompts that cause an erased model to regenerate the erased concept. These early studies show the need to evaluate erasure under adversarial prompts.

Later studies attempt to reactivate an erased concept through conditions other than text prompts. Memories of Forgotten Concepts \cite{rusanovsky2025memories} uses diffusion inversion to find latent seeds from which an erased model can reconstruct images containing the erased concept. A separate study \cite{george2025illusion} shows that an erased concept can reappear after the erased model is fine-tuned on data unrelated to that concept. M-ErasureBench \cite{M-ErasureBenc} brings text prompts, learned embeddings, and inverted latents into one multimodal benchmark, while RECALL \cite{RECALL} evaluates concept reactivation under adversarial image prompts. However, prior work has left the initial latent space underexplored. IVO extends concept reactivation attacks to this space and uses a semantic-anchoring loss that encourages consistency between the generated image and the target prompt. Table~\ref{tab:attack_comparison} summarizes the differences between IVO and the other attacks.

\begin{table}[t]
\centering

\caption{Compatibility of concept erasure methods across different generative model architectures. The columns show representative models under diffusion and flow matching. A \cmark~indicates that the method was evaluated on this model in the original paper, or that its official repository provides adaptation code. A \xmark~indicates that neither condition is met.}
\label{tab:model_compatibility}

\small
\setlength{\tabcolsep}{3.5pt}
\renewcommand{\arraystretch}{1.08}

\begin{tabular}{l c c c c c}

\toprule

\multirow{2}{*}{\textbf{Methods}}
& \multicolumn{3}{c}{\textbf{Diffusion}}
& \multicolumn{2}{c}{\textbf{Flow Matching}}
\\[-1pt]

\cmidrule(lr){2-4}
\cmidrule(lr){5-6}

& SDv1 & SDv2 & SDXL & SDv3 & FLUX \\

\midrule

ESD \cite{ESD} & \cmark & \xmark & \cmark & \xmark & \cmark \\

FMN \cite{FMN} & \xmark & \cmark & \xmark & \xmark & \xmark \\

SPM \cite{SPM} & \cmark & \cmark & \cmark & \xmark & \xmark \\

CPE \cite{CPE} & \cmark & \xmark & \xmark & \xmark & \xmark \\

AGE \cite{AGE} & \cmark & \cmark & \xmark & \xmark & \xmark \\

AdvUnlearn \cite{AdvUnlearn} & \cmark & \xmark & \xmark & \xmark & \xmark \\

STEREO \cite{stereo} & \cmark & \xmark & \xmark & \xmark & \xmark \\

AEGIS \cite{AEGIS} & \cmark & \cmark & \xmark & \xmark & \xmark \\

UCE \cite{UCE} & \cmark & \cmark & \cmark & \xmark & \cmark \\

RECE \cite{rece} & \cmark & \cmark & \xmark & \xmark & \xmark \\

SLD \cite{SLD} & \cmark & \cmark & \xmark & \xmark & \xmark \\

MACE \cite{MACE} & \cmark & \xmark & \xmark & \xmark & \xmark \\

SAeUron \cite{SAeUron} & \cmark & \xmark & \xmark & \xmark & \xmark \\

NeurErase \cite{NeurErase} & \cmark & \xmark & \cmark & \xmark & \xmark \\

ActErase \cite{ActErase} & \cmark & \xmark & \xmark & \xmark & \xmark \\

CASteer \cite{casteer} & \cmark & \cmark & \cmark & \xmark & \xmark \\

RAZOR \cite{RAZOR} & \cmark & \xmark & \xmark & \cmark & \xmark \\

EraseAnything \cite{eraseanything} & \xmark & \xmark & \xmark & \xmark & \cmark \\

EraseAnything++ \cite{eraseanything++} & \xmark & \xmark & \xmark & \xmark & \cmark \\

HiRM \cite{HiRM} & \cmark & \xmark & \xmark & \xmark & \cmark \\

SalUn \cite{SalUn} & \cmark & \xmark & \xmark & \xmark & \xmark \\

ConceptPrune \cite{ConceptPrune} & \cmark & \xmark & \xmark & \xmark & \xmark \\

\specialrule{\lightrulewidth}{0pt}{0pt}

\textbf{CeFlune (ours)} & \cmark & \cmark & \cmark & \cmark & \cmark \\

\specialrule{\heavyrulewidth}{0pt}{0pt}
\end{tabular}
\end{table}

\begin{table}[h]
\centering
\caption{Comparison of concept reactivation attacks. RECALL optimizes an image latent $\mathbf{z}_0$ encoded by a variational autoencoder (VAE), whereas IVO optimizes $\mathbf{z}_T$, the initial latent used to start image generation. In the Semantic Consistency column, \cmark~indicates that the attack design explicitly considers alignment with the target prompt, and \xmark~indicates that it does not.}
\label{tab:attack_comparison}

\small
\setlength{\tabcolsep}{0.7mm}
\renewcommand{\arraystretch}{1.18}

\begin{tabular}{l c c c}

\toprule

\textbf{Method} & \textbf{Access} & \textbf{Attack Variable} & \makecell{\textbf{Semantic}\\\textbf{Consistency}} \\

\midrule
CI \cite{CI} & White-box & Text embedding & \ding{55} \\

Ring-A-Bell \cite{Ring-A-Bell} & Black-box & Prompt tokens & \ding{55} \\

P4D \cite{P4D} & White-box & Prompt tokens & \ding{55} \\

UnlearnDiffAtk \cite{UnlearnDiffAtk} & White-box & Prefix tokens & \ding{55} \\

RECALL \cite{RECALL} & White-box & Image latent & \ding{51} \\

\midrule[0.4pt]

\textbf{IVO (ours)} & White-box & Initial latent & \ding{51} \\

\bottomrule
\end{tabular}
\end{table}
\subsection{Concept Localization} \label{sec:Concept Localization}

Existing studies identify model components associated with a concept, but these components vary in granularity. ConceptPrune \cite{ConceptPrune} and Forget-It-All \cite{Forget-It-All} both score individual weights in the second linear layer of the feed-forward network. They compare the contributions of these weights with and without the concept in the prompt, then prune the selected weights. NeurErase \cite{NeurErase} identifies concept-related neurons in cross-attention value projections and deactivates them during early denoising steps. At a coarser structural scale, RAZOR \cite{RAZOR} uses gradients from the forget and retain concept sets to rank text encoder modules for selective updating. A recent study \cite{zarei2026localizing} localizes concepts to specific diffusion transformer blocks and fine-tunes those blocks for erasure. Alongside these structural choices, localization can also choose particular stages of the generation process. For example, SDErasure \cite{SDErasure} selects the generation steps that are most critical for a concept to appear in the generated image.

Other studies use intermediate model outputs (activations) to locate concept-related representations rather than model components. SAeUron \cite{SAeUron} and other methods based on sparse autoencoders (SAEs) \cite{cassano2025saemnesia,he2025single,shi2026orthoeraser} learn sparse features from these outputs and select those associated with the target concept. ActErase \cite{ActErase} derives concept-related activation patterns by comparing activations under target and unrelated prompts, while PURE \cite{PURE} constructs concept subspaces from cross-attention activations.

Prior work varies in the structural granularity, temporal scope, and representation used for localization. However, many of these studies emphasize localization but pay less attention to effective intervention based on the localization results. CeFlune simulates brain functional localization in text-to-image models to locate concept-responsive Unit Neurons, then fine-tunes their parameters rather than simply pruning them. Its prediction-space loss guides model predictions toward concept erasure while helping preserve general image generation capability. A task-to-rest response alignment loss also constrains the selected Unit Neurons by aligning their task-state responses with the original model’s resting-state responses.

\section{Preliminaries and Problem Definition} \label{sec:Preliminaries and Problem Definition}
This section begins with a unified description of diffusion and flow-matching text-to-image models, providing a general generative framework for both IVO and CeFlune. It then formulates the objectives of the two tasks studied in this work: concept reactivation attack and concept erasure.

\subsection{Text-to-Image Generative Models} \label{sec:Text-to-Image Generative Models}
Latent diffusion and flow-matching text-to-image models generate images through repeated condition-dependent predictions. Although they construct and update intermediate states differently, both can be described using a common notation. Let $\mathbf{z}_0$ denote the clean latent representation of an image, $t$ a diffusion timestep or flow time, and $\boldsymbol{\epsilon}\sim\mathcal{N}(0,I)$ a Gaussian noise sample. The state at time $t$ is
\begin{equation}
\mathbf{z}_t = \alpha_t\mathbf{z}_0 + \sigma_t\boldsymbol{\epsilon} .
\label{eq:latent-state}
\end{equation}

This state construction applies to both latent diffusion models and flow-matching models. For latent diffusion \cite{croitoru2023diffusion}, $\alpha_t=\sqrt{\bar\alpha_t}$ and $\sigma_t=\sqrt{1-\bar\alpha_t}$, where $\bar\alpha_t$ denotes the value of the diffusion noise schedule at timestep $t$. For the linear flow-matching path \cite{lipman2022flow,esser2024scaling}, $\sigma_t\in[0,1]$ is the interpolation coefficient of the noise term, and $\alpha_t=1-\sigma_t$.

The state $\mathbf z_t$ serves as the model input at time $t$. Given $\mathbf z_t$ and a text condition $q$, the conditional predictor $F_\theta$, parameterized by $\theta$, produces
\begin{equation}
\mathbf p_\theta(\mathbf z_t,q,t) = F_\theta(\mathbf z_t,q,t).
\end{equation}

For an $\epsilon$-prediction latent diffusion model \cite{croitoru2023diffusion}, $\mathbf p_\theta=\boldsymbol\epsilon_\theta$, where $\boldsymbol\epsilon_\theta$ is the model prediction of the sampled noise $\boldsymbol\epsilon$. For a flow-matching model \cite{lipman2022flow,esser2024scaling}, $\mathbf p_\theta=\mathbf v_\theta$ is the velocity field prediction.

Complete generation chains these single step predictions. Starting from the initial latent noise $\mathbf z_T$, the sampler repeatedly evaluates $F_\theta$ to update the intermediate state toward a clean latent, which is then decoded into an image:
\begin{equation}
\mathbf x = G_\theta(\mathbf z_T,q) = \operatorname{Decode}\!\left(
\operatorname{Sample}(F_\theta,\mathbf z_T,q)
\right) ,
\end{equation}
where $G_\theta$ denotes the complete generation process.

\subsection{Concept Reactivation Attack Objectives} \label{sec:Concept Reactivation Attack Objectives}
\hspace*{\parindent}\textbf{Concept reactivation.} Let $C_{\mathrm{f}}$ denote the set of target concepts, and let $\mathcal{A}$ denote the set of attack methods. The erased model has parameters $\tilde{\theta}$, which remain fixed during the attack. For $c\in C_{\mathrm{f}}$ and $\pi\in\mathcal{A}$, let $\mathbf{x}_{c}^{\pi}$ denote the output image produced by attack method $\pi$ when applied to target concept $c$. $s_c(\mathbf{x})\in[0,1]$ measures how strongly image $\mathbf{x}$ expresses $c$. The concept reactivation score is
\begin{equation}
A_\pi(\tilde{\theta};C_{\mathrm{f}}) = \mathbb{E}_{c\in C_{\mathrm{f}}}
\left[s_c\!\left(\mathbf{x}_{c}^{\pi}\right)\right].    
\end{equation}

When $s_c$ is a binary concept detector, $A_\pi$ is reported as attack success rate (ASR). A higher ASR indicates more successful concept reactivation.

\textbf{Semantic consistency.} ASR alone is insufficient to confirm a successful attack, because a high ASR can result from fooling the concept detector. The attack output should also remain as semantically consistent as possible with the target prompt. Let $d_{\mathrm{sem}}(\mathbf{x},q_c)$ measure the semantic discrepancy between image $\mathbf{x}$ and target prompt $q_c$. The attack output is required to satisfy
\begin{equation}
d_{\mathrm{sem}}\!\left(\mathbf{x}_{c}^{\pi},q_c\right)\leq \varepsilon_{\mathrm{sem}},
\end{equation}
where $\varepsilon_{\mathrm{sem}}$ is the semantic discrepancy threshold.

Throughout this work, we use CLIPScore (CLIP) \cite{hessel2021clipscore} as the default measure of semantic consistency, with higher values indicating better text–image alignment.

\subsection{Concept Erasure Objectives} \label{sec:Concept-Erasure Objectives}
\hspace*{\parindent}\textbf{Target concept erasure.} In the absence of an attack, target concept erasure is quantified by the erasure rate (ER), defined as
\begin{equation}
\mathrm{ER}(\tilde{\theta};C_{\mathrm{f}}) = 1 - \mathbb{E}_{c\in C_{\mathrm{f}}}\left[s_c\!\left(G_{\tilde{\theta}}(\mathbf{z}_T,q_c)\right)\right].
\end{equation}

When $s_c$ is a binary concept detector, $\mathrm{ER}$ is the fraction of generated images in which the target concept is not detected. A higher ER indicates stronger erasure.

\textbf{Generative performance preservation.} This objective comprises related concept preservation and unrelated concept preservation. The former limits the effect of erasure on concepts that are semantically or visually similar to the target concept $c$. The latter limits its effect on concepts unrelated to $c$. Let $C_{\mathrm{k}}$ denote either set of concepts to retain, and let $d(\cdot,\cdot)$ measure the difference between images generated by the erased model $\tilde{\theta}$ and the original model $\theta$ from the same input. The difference is formulated as
\begin{equation}
D(\tilde{\theta},\theta;C_{\mathrm{k}}) = \mathbb{E}_{c_{\mathrm{k}}\in C_{\mathrm{k}}} \left[d\!\left( G_{\tilde{\theta}}(\mathbf{z}_T,q_{c_{\mathrm{k}}}), G_{\theta}(\mathbf{z}_T,q_{c_{\mathrm{k}}})\right)\right]. 
\end{equation}

In this work, we quantify $d(\cdot,\cdot)$ using ER and CLIP. We compute both metrics for the erased and original models on the retained concept set and compare their values. A smaller increase in ER and a smaller decrease in CLIP relative to the original model indicate better preservation.

\textbf{Attack resistance.} The worst case measured reactivation score over $\mathcal{A}$ is
\begin{equation}
R(\tilde{\theta};\mathcal{A}) = \max_{\pi\in\mathcal{A}} A_\pi(\tilde{\theta};C_{\mathrm{f}}).
\end{equation}

Attack resistance is evaluated using ASR, where a lower value indicates stronger resistance.
We reformulate the three objectives as a constrained optimization problem:
\begin{equation}
\begin{gathered}
\theta^\star \in \operatorname*{arg\,max}_{\tilde{\theta}} \mathrm{ER}(\tilde{\theta};C_{\mathrm{f}}) \\
\text{s.t.}\quad D(\tilde{\theta},\theta;C_{\mathrm{k}}) \leq \varepsilon_{\mathrm{keep}} \quad \text{and} \quad R(\tilde{\theta};\mathcal{A}) \leq \varepsilon_{\mathrm{atk}},
\end{gathered}   
\end{equation}
where $\varepsilon_{\mathrm{keep}}$ and $\varepsilon_{\mathrm{atk}}$ denote the upper bounds imposed on $D$ and $R$, respectively.

\begin{figure}[h]
    \centering
    \includegraphics[width=\columnwidth]{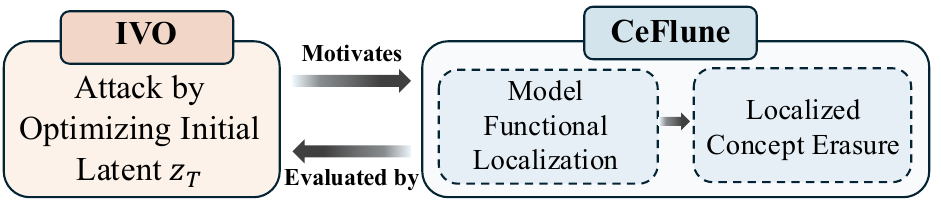}
    \caption{The relationship between IVO and CeFlune. IVO extends concept reactivation attacks to the initial latent space, motivating CeFlune's model functional localization and localized concept erasure. IVO then evaluates the erased model’s attack resistance.}
    \label{fig:method_overview}
\end{figure}

\section{Proposed Methods} \label{sec:Proposed Method}

\subsection{Overview} \label{sec:Overview}
Fig.~\ref{fig:method_overview} summarizes the relationship between IVO and CeFlune. IVO extends the evaluation of concept reactivation to the initial latent space (Section~\ref{sec:IVO: Initial Latent Variable Optimization}). This new risk motivates CeFlune, which first localizes concept-responsive Unit Neurons through task–rest response comparison (Section~\ref{sec:CeFlune: Model Functional Localization}) and then updates the selected Unit Neurons for concept erasure (Section~\ref{sec:CeFlune: Localized Concept Erasure}). IVO is subsequently used to evaluate the resulting model’s attack resistance.

\subsection{IVO: Initial Latent Variable Optimization} \label{sec:IVO: Initial Latent Variable Optimization}

\begin{figure}[t]
    \centering
    \includegraphics[width=\columnwidth]{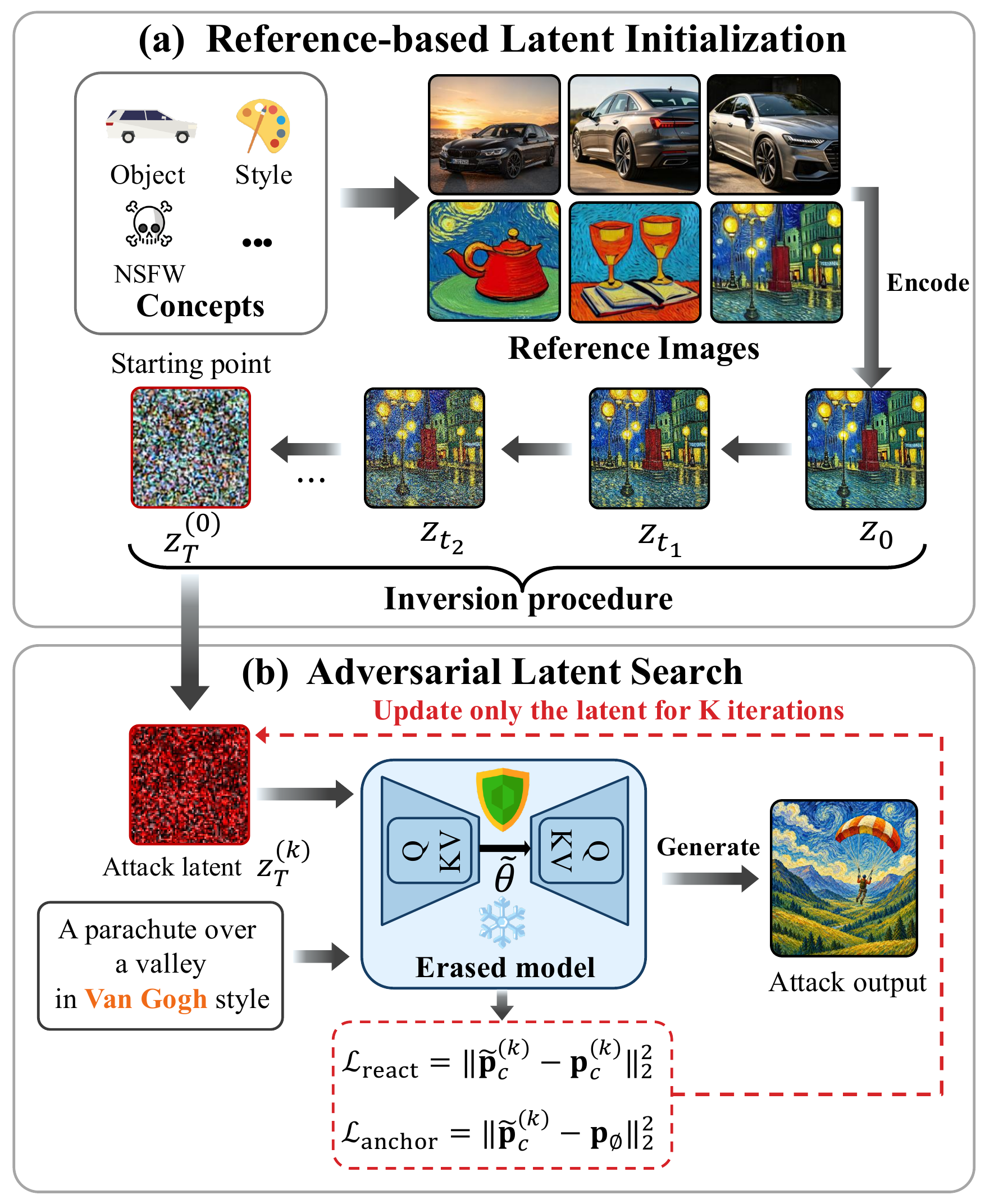}
    \caption{Initial latent variable optimization (IVO). (a) A reference image is encoded into $\mathbf{z}_0$, which is inverted to obtain the initial latent $\mathbf{z}_T^{(0)}$. (b) IVO optimizes this latent using concept-reactivation and semantic-anchoring losses. The erased model then generates an image using the optimized latent $\mathbf{z}_T^{(K)}$ and target prompt $q_c$.}
    \label{fig:ivo-framework}
\end{figure}

For a target concept $c\in C_{\mathrm{f}}$ and its target prompt $q_c$, IVO uses an optimized initial latent variable to attack the erased model. As shown in Fig.~\ref{fig:ivo-framework}, IVO initializes the latent variable from a reference image, and optimizes it using two losses. The concept-reactivation loss encourages the erased model to generate an image containing the target concept. The semantic-anchoring loss encourages the generated image to remain semantically consistent with the target prompt $q_c$.

\subsubsection{Reference-Based Latent Initialization} \label{sec:Reference-Based Latent Initialization}

Given a reference image $\mathbf{x}_{\mathrm{ref}}$ containing the target concept $c$ and its textual description $q$ generated by a multimodal model, we first encode the image into a clean latent $\mathbf{z}_0$. An inversion procedure then approximately reverses the generation process of the original model $F_{\theta}$, mapping $\mathbf{z}_0$ to an initial latent:
\begin{equation}
\mathbf{z}_T^{(0)}
=
\operatorname{Inv}_{\theta}(\mathbf{z}_0,q)
\quad \text{s.t.} \quad
G_{\theta}(\mathbf{z}_T^{(0)},q)
\approx
\mathbf{x}_{\mathrm{ref}},
\end{equation}
where $\operatorname{Inv}_{\theta}$ denotes the inversion procedure and $\mathbf{z}_T^{(0)}$ serves as the starting point for IVO optimization.

\subsubsection{Adversarial Latent Search} \label{sec:Prediction-Matching Attack Objective}
Section~\ref{sec:Concept Reactivation Attack Objectives} defines the attack objective as inducing the erased model to generate an image that expresses concept $c$ and remains semantically consistent with $q_c$. IVO optimizes the initial latent toward this objective using a concept-reactivation loss and a semantic-anchoring loss.

At attack iteration $k$, the original and erased models receive the same attack latent $\mathbf z_T^{(k)}$, target prompt $q_c$, and time $t$. The original model produces $\mathbf p_c^{(k)}=F_{\theta}(\mathbf z_T^{(k)},q_c,t)$, and the erased model produces $\tilde{\mathbf p}_c^{(k)}=F_{\tilde{\theta}}(\mathbf z_T^{(k)},q_c,t)$. The concept-reactivation loss penalizes the squared $L_2$ distance between $\tilde{\mathbf p}_c^{(k)}$ and $\mathbf p_c^{(k)}$.

At the same time $t$, the original model also produces an empty text prediction $\mathbf p_{\varnothing}=F_{\theta}(\mathbf z_T^{(0)},\varnothing,t)$. Without direct guidance from $q_c$, this prediction reflects semantic prior carried by $\mathbf z_T^{(0)}$. The semantic-anchoring loss penalizes the squared $L_2$ distance between $\tilde{\mathbf p}_c^{(k)}$ and $\mathbf p_{\varnothing}$. We define the IVO optimization objective as the sum of these two losses:
\begin{equation}
\mathcal{L}_{\mathrm{IVO}}(\mathbf{z}_T^{(k)}) = \underbrace{\left\|\tilde{\mathbf{p}}_c^{(k)}-\mathbf p_c^{(k)}\right\|_2^2}_{\mathcal{L}_{\mathrm{react}}} + \underbrace{\left\|\tilde{\mathbf{p}}_c^{(k)}-\mathbf{p}_{\varnothing}\right\|_2^2}_{\mathcal{L}_{\mathrm{anchor}}}.
\label{eq:ivo-objective}
\end{equation}

IVO updates the initial latent $\mathbf z_T^{(0)}$ over $K$ iterations to minimize $\mathcal{L}_{\mathrm{IVO}}$. The erased model then generates an image using the optimized latent $\mathbf z_T^{(K)}$:
\begin{equation}
\mathbf{x}_c^{\mathrm{IVO}}=G_{\tilde{\theta}}(\mathbf{z}_T^{(K)},q_c).
\end{equation}

\begin{figure}[t]
    \centering
    \includegraphics[width=0.98\linewidth]{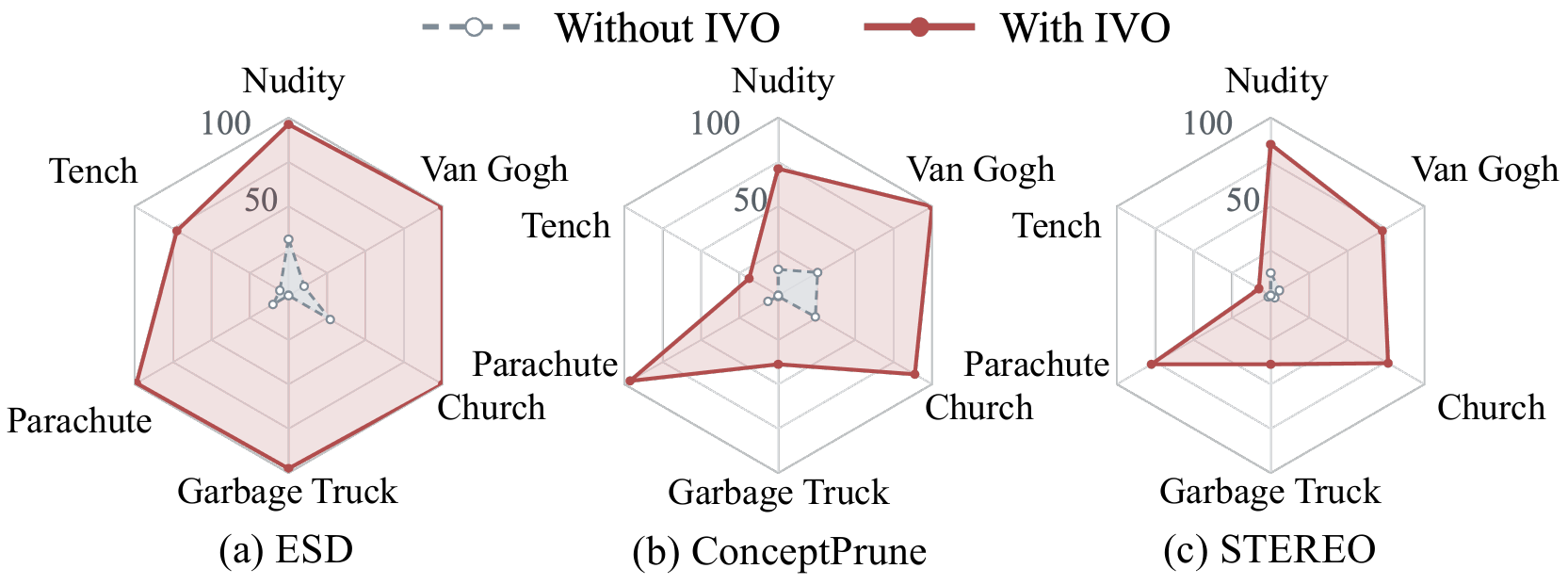}
    \caption{Concept reactivation by IVO in SDv1.4 models erased by ESD, ConceptPrune, and STEREO. The radial values show the percentage of generated images in which each of the six core concepts is detected.}
    \label{fig:ivo-radar}
\end{figure}

Fig.~\ref{fig:ivo-radar} offers a brief illustration of IVO’s ability to reactivate erased concepts. Section~\ref{sec:Reactivation under Attacks} extends this evaluation to nine erasure methods and five attacks.

\subsection{CeFlune: Model Functional Localization} \label{sec:CeFlune: Model Functional Localization}

\begin{figure*}[!t]
    \centering
    \includegraphics[width=\textwidth]{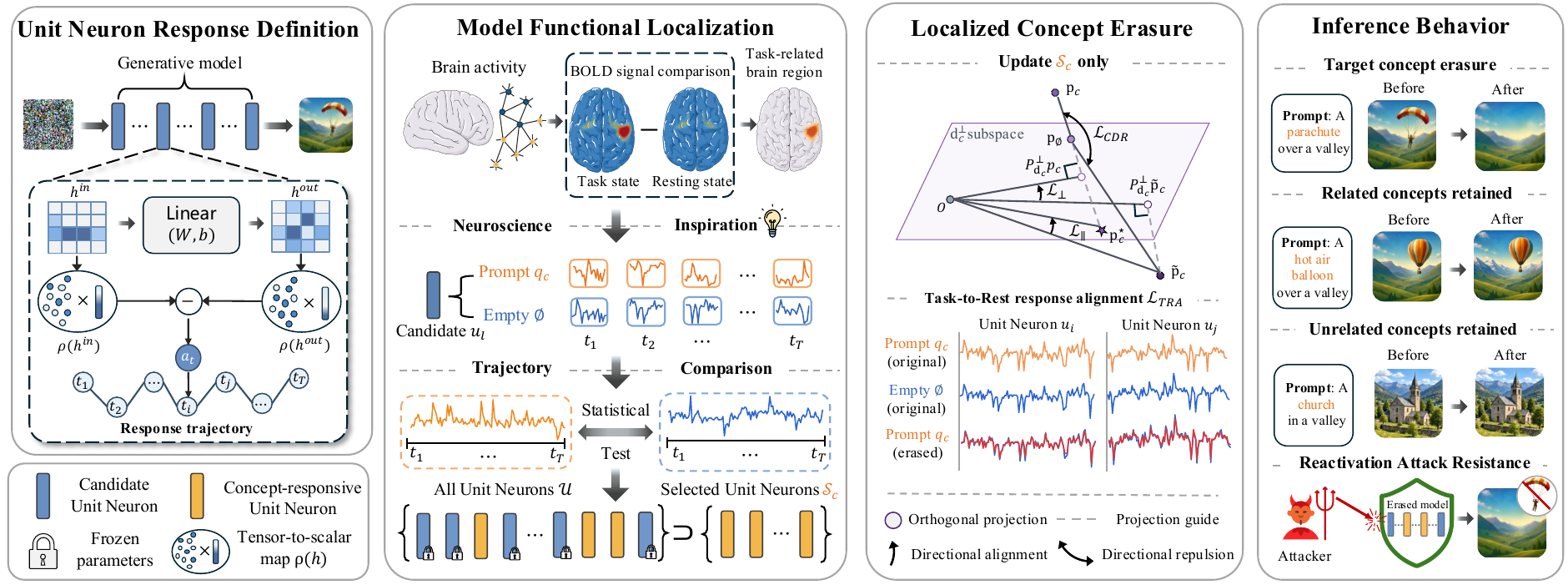}
    \caption{CeFlune framework. ``Unit Neuron Response Definition'' derives a Unit Neuron's response from its input and output tensors. ``Model Functional Localization'' records responses over time and compares the task-state and resting-state trajectories to select concept-responsive Unit Neurons, mirroring the task–rest BOLD comparison. ``Localized Concept Erasure'' depicts the prediction-space constraints and alignment of the selected Unit Neurons’ task-state responses with the original model’s resting-state responses. ``Inference Behavior'' provides examples of target concept erasure, preservation of other concepts, and resistance to reactivation attacks.}
    \label{fig:ceflune-framework}
\end{figure*}

Fig.~\ref{fig:ceflune-framework} provides an overview of CeFlune. This section presents CeFlune’s simulation of brain functional localization in text-to-image models, which we call model functional localization. Section~\ref{sec:From Brain to Model} first introduces brain functional localization in fMRI and, by analogy, defines the Unit Neuron and a general form for measuring its activity in text-to-image models. Sections~\ref{sec:Unit Neuron Response Measurement} and \ref{sec:Concept-Responsive Unit Neuron Selection} then specify the statistic used to measure that activity and the rule for selecting concept-responsive Unit Neurons, respectively.

\subsubsection{From Brain to Model} \label{sec:From Brain to Model}

Brain functional localization seeks to identify brain regions associated with a particular task. Task-based functional magnetic resonance imaging (fMRI) is widely used for this purpose \cite{logothetis2001neurophysiological}. It measures changes in the blood-oxygen-level-dependent (BOLD) signal at the voxel level, with each voxel representing a small volume of brain tissue containing many neurons \cite{olman2011high}. The BOLD signal is recorded while a participant performs a prescribed task (task state). It is also recorded while the participant rests without performing the task (resting state) \cite{friston1994statistical}. Generally, the participant’s brain remains active during the resting period, so the BOLD signal recorded in this period can serve as baseline brain activity. The task-period measurement is then compared with this baseline. The resulting difference is called the task-evoked response and is used to localize task-responsive voxels. The identified voxels delineate task-related brain regions.

CeFlune formulates an analogous process for text-to-image models by first defining what unit to localize and giving a general form for measuring its activity. Like a voxel containing many biological neurons, a linear module also comprises many conventional artificial neurons as understood in machine learning. Therefore, we define each linear module in $F_\theta$ as a Unit Neuron, which serves as the basic unit of measurement, localization, and intervention. Let $u_l$ denote the $l$-th Unit Neuron. Its input and output dimensions are $d_l^{\mathrm{in}}$ and $d_l^{\mathrm{out}}$. Its weight matrix and bias vector are $\mathbf{W}_l$ and $\mathbf{b}_l$. At time $t$ under text condition $q$, $\mathbf{h}_l^{\mathrm{in}}(t,q)$ and $\mathbf{h}_l^{\mathrm{out}}(t,q)$ denote its input and output tensors respectively. They satisfy
\begin{equation}
u_l: \mathbf{h}_l^{\mathrm{out}}(t,q) = u_l\!\left(\mathbf{h}_l^{\mathrm{in}}(t,q)\right) = \mathbf{W}_l\mathbf{h}_l^{\mathrm{in}}(t,q)+\mathbf{b}_l.
\end{equation}

To characterize the activity of $u_l$, we define its response $a_l(t,q)$ from input and output tensors. Let $\rho(\cdot)$ denote a function that maps a tensor to a scalar. The response is
\begin{equation}
a_l(t,q) = \rho\!\left(\mathbf{h}_l^{\mathrm{out}}(t,q)\right) - \rho\!\left(\mathbf{h}_l^{\mathrm{in}}(t,q)\right).
\label{eq:unit-neuron-response}
\end{equation}

We next construct two model states using different text conditions. A prompt $q_c$ expressing concept $c$ defines the task condition and induces the task state. Empty text $\varnothing$ defines the resting condition and induces the resting state. The corresponding Unit Neuron responses are $a_l(t,q_c)$ and $a_l(t,\varnothing)$ respectively. Just as the BOLD signal recorded during rest represents baseline brain activity in fMRI, $a_l(t,\varnothing)$ provides baseline Unit Neuron activity.

CeFlune compares task-state and resting-state Unit Neuron responses to obtain the concept-evoked response. This task–rest response comparison provides the basis for localizing concept-responsive Unit Neurons, which form the set $\mathcal{S}_c$ used for localized concept erasure. We provide Table~\ref{tab:fmri_CeFlune_correspondence} to compare key elements of brain functional localization and model functional localization.

\begin{table}[t]
\caption{Correspondence between functional localization in fMRI and CeFlune.}
\label{tab:fmri_CeFlune_correspondence}
\centering
\small

\setlength{\tabcolsep}{0.5mm}
\renewcommand{\arraystretch}{1.2}

\resizebox{\columnwidth}{!}{
\begin{tabular}{l c c}

\toprule

\textbf{Elements} & \textbf{fMRI} & \textbf{CeFlune} \\

\midrule

\textbf{Unit} & Voxel & Unit Neuron $u_l$ \\

\textbf{Measure} & BOLD signal & $a_l(t,q)$ \\

\textbf{Condition} & Task/Rest & $q_c$/$\varnothing$ \\

\textbf{Comparison} & Task vs. Rest & $q_c$ vs. $\varnothing$ \\

\textbf{Evoked response} & Task-evoked & Concept-evoked \\

\textbf{Responsive unit} & Task-responsive voxel & Concept-responsive $u_l$ \\

\midrule

\textbf{Result} & Task-related brain regions & Set $\mathcal{S}_c$ \\

\bottomrule

\end{tabular}}
\end{table}

\subsubsection{Unit Neuron Response Measurement} \label{sec:Unit Neuron Response Measurement}

Let $N(\mathbf{h})$ denote the number of elements in $\mathbf{h}$, and let $h_n$ denote the $n$-th element. For each $h_n$, CeFlune computes its normalized magnitude as $\hat h_n=|h_n|/(\lVert\mathbf{h}\rVert_2+\epsilon)$ and the corresponding binary mask as $m_n(\mathbf{h})=\mathbb{I}[\hat h_n>\epsilon]$, where $\epsilon>0$. Using these quantities, we define the localization statistic as:
\begin{equation}
\psi(\mathbf{h})= \left(\frac{1}{N(\mathbf{h})}\sum_n m_n(\mathbf{h})\right)\left(\sum_n \hat h_n m_n(\mathbf{h})\right),
\end{equation}
where the first factor is the fraction of retained elements and the second is the sum of their normalized magnitudes.

For each random seed $r$, substituting $\psi(\cdot)$ for the generic tensor-to-scalar mapping $\rho(\cdot)$ in Eq.~\ref{eq:unit-neuron-response} gives:
\begin{equation}
a_{l,r}(t,q) = \psi\!\left(\mathbf{h}^{\mathrm{out}}_{l,r}(t,q)\right) - \psi\!\left(\mathbf{h}^{\mathrm{in}}_{l,r}(t,q)\right).
\label{eq:localization-response}
\end{equation}

However, $a_{l,r}(t,q)$ only measures the response of a Unit Neuron at a given time. We collect its responses over $t=1,\ldots,T$ to form a full response trajectory $\mathbf{a}_{l,r}(q)=[a_{l,r}(1,q),\ldots,a_{l,r}(T,q)]^{\top}$, which corresponds to the BOLD signal recorded over time in fMRI.

\subsubsection{Concept-Responsive Unit Neuron Selection} \label{sec:Concept-Responsive Unit Neuron Selection}

Let $\mathcal{U}=\{u_l\}_{l=1}^{L}$ denote the candidate Unit Neurons enumerated from the original model $F_{\theta}$. For each $u_l\in\mathcal{U}$ and text condition $q$, CeFlune averages the response trajectories over $N$ random seeds $r\in\{1,\ldots,N\}$:
\begin{equation}
\bar{\mathbf{a}}_l(q)
=
\frac{1}{N}
\sum_{r=1}^{N}
\mathbf{a}_{l,r}(q).
\end{equation}

Setting $q=q_c$ and $q=\varnothing$ gives the mean task-state trajectory $\bar{\mathbf{a}}_l(q_c)$ and the mean resting-state trajectory $\bar{\mathbf{a}}_l(\varnothing)$, respectively. CeFlune uses a trajectory-level test $\mathcal{T}$ to assess their correlation. At threshold $\tau$, the test returns the localization decision
\begin{equation}
\delta_l^c
=
\mathcal{T}\!\left(
\bar{\mathbf{a}}_l(q_c),
\bar{\mathbf{a}}_l(\varnothing);
\tau
\right)
\in\{0,1\},
\end{equation}
where $\delta_l^c=0$ indicates that the two trajectories are significantly correlated, and $\delta_l^c=1$ indicates that they are not. Unit Neurons with $\delta_l^c=1$ are concept-responsive Unit Neurons and form the selected set $\mathcal{S}_c=\{u_l\in\mathcal{U}\mid \delta_l^c=1\}$.

\subsection{CeFlune: Localized Concept Erasure} \label{sec:CeFlune: Localized Concept Erasure}
Model functional localization determines where CeFlune intervenes by identifying the concept-responsive Unit Neurons. We next describe how CeFlune updates these Unit Neurons by jointly optimizing three prediction-space losses and a task-to-rest response alignment loss. The prediction-space losses guide the erased model's prediction under $q_c$ away from target concept direction while preserving components orthogonal to that direction. The alignment loss anchors each selected Unit Neuron's response under $q_c$ to the original model's response under empty text $\varnothing$.

\textbf{Preparation for erasure optimization.} Let $\mathcal{D}_{\mathrm{aux}}$ denote an auxiliary image distribution unrelated to the target concept $c$. Let $\theta^{(j)}$ denote the model parameters at erasure iteration $j$, with $\theta^{(0)}=\theta$. We sample $\mathbf{x}\sim\mathcal{D}_{\mathrm{aux}}$ and encode it as $\mathbf{z}_0$. After sampling time $t$, we construct the corresponding noised latent $\mathbf{z}_t$ from $\mathbf{z}_0$ using the diffusion or flow-matching state equation in Eq.~\ref{eq:latent-state}. At each erasure optimization iteration, CeFlune feeds the same noised latent $\mathbf z_t$ and time $t$ to the original and erased models. The frozen original model provides the task-state prediction $\mathbf{p}_c=F_{\theta}(\mathbf{z}_t,q_c,t)$ and the resting-state prediction $\mathbf{p}_{\varnothing}=F_{\theta}(\mathbf{z}_t,\varnothing,t)$. These predictions serve as references for constraining the erased model's task-state prediction $\tilde{\mathbf{p}}_c=F_{\theta^{(j)}}(\mathbf{z}_t,q_c,t)$.

\subsubsection{Prediction-Space Erasure and Preservation} \label{sec:Prediction-Space Erasure and Preservation}
For prediction tensors $\mathbf{u}$ and $\mathbf{v}$, we define their cosine similarity as
\begin{equation}
\operatorname{sim}_{\cos}(\mathbf{u},\mathbf{v})
= \frac{\langle\operatorname{vec}(\mathbf{u}),\operatorname{vec}(\mathbf{v})\rangle} {\|\operatorname{vec}(\mathbf{u})\|_2\|\operatorname{vec}(\mathbf{v})\|_2},  
\end{equation}
where $\operatorname{vec}(\cdot)$ flattens a tensor into a vector.

\textbf{Negative guidance alignment.} Let $\Delta\mathbf{p}_c=\mathbf{p}_c-\mathbf{p}_{\varnothing}$ denote the difference between the original model's task-state and resting-state predictions. Using the negative guidance principle of ESD \cite{ESD}, we construct $\mathbf{p}_c^{\star}=\mathbf{p}_c-\eta\Delta\mathbf{p}_c$, where $\eta$ controls the guidance strength. When $\eta=1$, $\mathbf{p}_c^{\star}$ equals the resting-state prediction $\mathbf{p}_{\varnothing}$. When $\eta>1$, $\mathbf{p}_c^{\star}$ moves past $\mathbf{p}_{\varnothing}$ in the direction opposite to $\Delta\mathbf p_c$, farther away from the task-state prediction $\mathbf{p}_c$. We align the erased model's task-state prediction $\tilde{\mathbf{p}}_c$ with $\mathbf{p}_c^{\star}$ by minimizing the following loss:
\begin{equation}
\mathcal{L}_{\parallel} = 1-\operatorname{sim}_{\cos} \left(\tilde{\mathbf{p}}_c,\mathbf{p}_c^{\star}\right).
\end{equation}

\textbf{Orthogonal projection alignment.} We represent the target concept direction as $\mathbf{d}_c=\operatorname{vec}(\Delta\mathbf{p}_c)/\|\operatorname{vec}(\Delta\mathbf{p}_c)\|_2$. For any prediction $\mathbf y$, the orthogonal projection $P_{\mathbf{d}_c}^{\perp}\mathbf{y}=\operatorname{vec}(\mathbf{y})-\langle\operatorname{vec}(\mathbf{y}),\mathbf{d}_c\rangle\mathbf{d}_c$ removes its component along $\mathbf{d}_c$. 

The shift from $\mathbf{p}_c$ to $\mathbf{p}_c^{\star}$ is $-\eta\Delta\mathbf{p}_c$, which lies along $\mathbf{d}_c$. This means that $\mathbf{p}_c$ and $\mathbf{p}_c^{\star}$ have the same orthogonal projection, $P_{\mathbf{d}_c}^{\perp}\mathbf{p}_c^{\star}=P_{\mathbf{d}_c}^{\perp}\mathbf{p}_c$. The erased model's prediction $\tilde{\mathbf p}_c$ should also maintain this relationship with $\mathbf p_c$ under orthogonal projection. To preserve this relationship explicitly, $\mathcal L_{\perp}$ aligns the orthogonal projection of $\tilde{\mathbf p}_c$ with that of $\mathbf p_c$:
\begin{equation}
\mathcal{L}_{\perp}
=
1-\operatorname{sim}_{\cos}
\left(
P_{\mathbf{d}_c}^{\perp}\tilde{\mathbf{p}}_c,
P_{\mathbf{d}_c}^{\perp}\mathbf{p}_c
\right).
\end{equation}

This constraint limits changes in the erased model's prediction outside $\mathbf{d}_c$, helping preserve general image generation capability.

\textbf{Concept direction repulsion.} Even if $\mathcal L_{\parallel}$ reaches its minimum value of zero, this loss constrains only the direction of $\tilde{\mathbf p}_c$, allowing $\tilde{\mathbf p}_c=\alpha\mathbf p_c^\star$ for any $\alpha>0$. In other words, the two predictions point in the same direction from the origin, but their directions may differ relative to the resting-state prediction. To see this, we write:
\begin{equation}
\begin{aligned}
\tilde{\mathbf p}_c-\mathbf p_{\varnothing}
&=\alpha\mathbf p_c^\star-\mathbf p_{\varnothing}\\
&=(\alpha-1)\mathbf p_{\varnothing}
+\alpha(1-\eta)\Delta\mathbf p_c.
\end{aligned} 
\end{equation}

If $\alpha\neq1$, the term $(\alpha-1)\mathbf p_{\varnothing}$ may cause $\tilde{\mathbf p}_c-\mathbf p_{\varnothing}$ to align with $\Delta\mathbf p_c$. We minimize the cosine similarity between $\tilde{\mathbf p}_c-\mathbf p_{\varnothing}$ and $\Delta\mathbf p_c$ to reduce the risk of concept reactivation:
\begin{equation}
\mathcal{L}_{\mathrm{CDR}} = \operatorname{sim}_{\cos}\left(\tilde{\mathbf{p}}_{c}-\mathbf{p}_{\varnothing}, \mathbf{p}_{c}-\mathbf{p}_{\varnothing} \right).
\end{equation}

With $\lambda_{\perp}$ and $\lambda_{\mathrm{CDR}}$ weighting $\mathcal L_{\perp}$ and $\mathcal L_{\mathrm{CDR}}$, respectively, the prediction-space loss is:
\begin{equation}
\mathcal{L}_{\mathrm{pred}} = \mathcal{L}_{\parallel} + \lambda_{\perp}\mathcal{L}_{\perp} + \lambda_{\mathrm{CDR}}\mathcal{L}_{\mathrm{CDR}}  .
\end{equation}

\subsubsection{Task-to-Rest Response Alignment} \label{sec:Task-to-Rest Response Alignment}
The prediction-space loss constrains the erased model's task-state prediction but does not directly control the responses of the selected Unit Neurons. CeFlune additionally aligns the erased model's task-state response with the frozen original model's resting-state response for each $u_l\in\mathcal S_c$.

Eq.~\ref{eq:localization-response} constructs Unit Neuron responses using the localization statistic $\psi$. For response alignment, however, using $\psi$ directly could make each selected Unit Neuron's contribution depend partly on the number of elements in its input and output tensors. To reduce this dependence, we define a size-normalized form of $\psi$: $\phi(\mathbf h)=\psi(\mathbf h)/\sqrt{N(\mathbf h)}$. Substituting $\phi$ for $\rho$ in Eq.~\ref{eq:unit-neuron-response} gives the original model's resting-state response and the erased model's task-state response:
\begin{equation}
\left\{
\begin{aligned}
a_l(t,\varnothing) &= \phi\!\left(\mathbf{h}^{\mathrm{out}}_l(t,\varnothing)\right)-
\phi\!\left(\mathbf{h}^{\mathrm{in}}_l(t,\varnothing)\right)\\
\tilde a_l(t,q_c)&= \phi\!\left(\tilde{\mathbf{h}}^{\mathrm{out}}_l(t,q_c)\right)-
\phi\!\left(\tilde{\mathbf{h}}^{\mathrm{in}}_l(t,q_c)\right).
\end{aligned}
\right. 
\end{equation}

The task-to-rest response alignment (TRA) loss sums the squared difference between these responses over the concept-responsive Unit Neurons:
\begin{equation}
\mathcal{L}_{\mathrm{TRA}} = \sum_{u_l\in\mathcal{S}_c} \left(\tilde a_l(t,q_c)-a_l(t,\varnothing)\right)^2.
\end{equation}

\begin{figure}[!t]
\centering
\small
\begin{minipage}{\columnwidth}
\refstepcounter{CeFlunealgorithm}\label{alg:CeFlune}
\noindent\rule{\linewidth}{0.6pt}\\
\textbf{Algorithm~\theCeFlunealgorithm: CeFlune}\\[-0.42em]
\rule{\linewidth}{0.3pt}
\begin{algorithmic}[1]
\setlength{\itemsep}{2pt}
\Statex \textbf{Input:} frozen original model $F_{\theta}$, target prompt $q_c$
\Statex \textbf{Output:} erased model $F_{\tilde{\theta}}$
\Statex
\Statex \textbf{Phase I: Model Functional Localization}
\State Enumerate all candidate Unit Neurons $\mathcal U$ in $F_{\theta}$
\For{$r=1,\ldots,N$}
\ForAll{$u_l\in\mathcal U$}
\State Compute the task-state response trajectory $\mathbf a_{l,r}(q_c)$
\State Compute the resting-state response trajectory $\mathbf a_{l,r}(\varnothing)$
\EndFor
\EndFor
\ForAll{$u_l\in\mathcal U$}
\State $\bar{\mathbf a}_l(q_c)\gets N^{-1}\sum_{r=1}^{N}\mathbf a_{l,r}(q_c)$
\State $\bar{\mathbf a}_l(\varnothing)\gets N^{-1}\sum_{r=1}^{N}\mathbf a_{l,r}(\varnothing)$
\State $\delta_l^c\gets\mathcal T\bigl(\bar{\mathbf a}_l(q_c),\bar{\mathbf a}_l(\varnothing);\tau\bigr)$
\EndFor
\State $\mathcal S_c\gets\{u_l\in\mathcal U\mid\delta_l^c=1\}$
\Statex
\Statex \textbf{Phase II: Localized Concept Erasure}
\State Initialize $\theta^{(0)}\gets\theta$ and freeze parameters outside $\mathcal S_c$
\For{$j=0,\ldots,J-1$}
\State Sample $\mathbf x\sim\mathcal D_{\mathrm{aux}}$ and time $t$
\State Encode $\mathbf x$ as $\mathbf z_0$ and construct $\mathbf z_t$
\State $\mathbf p_c\gets F_{\theta}(\mathbf z_t,q_c,t)$ and $\mathbf p_{\varnothing}\gets F_{\theta}(\mathbf z_t,\varnothing,t)$
\State $\Delta\mathbf p_c\gets\mathbf p_c-\mathbf p_{\varnothing}$, then $\mathbf p_c^\star\gets\mathbf p_c-\eta\Delta\mathbf p_c$
\State $\tilde{\mathbf p}_c\gets F_{\theta^{(j)}}(\mathbf z_t,q_c,t)$
\State $\mathbf{d}_c \gets \operatorname{vec}(\Delta\mathbf{p}_c)/\|\operatorname{vec}(\Delta\mathbf{p}_c)\|_2$
\ForAll{$u_l\in\mathcal S_c$}
\State Compute $a_l(t,\varnothing)$ for $F_{\theta}$ and $\tilde a_l(t,q_c)$ for $F_{\theta^{(j)}}$
\EndFor
\State $\mathcal L_{\parallel}\gets1-\operatorname{sim}_{\cos}(\tilde{\mathbf p}_c,\mathbf p_c^\star)$
\State $\mathcal L_{\perp}\gets1-\operatorname{sim}_{\cos}(P_{\mathbf d_c}^{\perp}\tilde{\mathbf p}_c,P_{\mathbf d_c}^{\perp}\mathbf p_c)$
\State $\mathcal L_{\mathrm{CDR}}\gets\operatorname{sim}_{\cos}(\tilde{\mathbf p}_c-\mathbf p_{\varnothing},\Delta\mathbf p_c)$
\State $\mathcal L_{\mathrm{TRA}}\gets
\sum_{u_l\in\mathcal S_c}\bigl(\tilde a_l(t,q_c)-a_l(t,\varnothing)\bigr)^2$
\State $\mathcal L_{\mathrm{total}}\gets(\mathcal L_{\parallel}
+\lambda_{\perp}\mathcal L_{\perp}+\lambda_{\mathrm{CDR}}\mathcal L_{\mathrm{CDR}})
+\lambda_{\mathrm{TRA}}\mathcal L_{\mathrm{TRA}}$
\State $\theta^{(j+1)} \gets \theta^{(j)} - \gamma M_{\mathcal{S}_c} \odot \nabla_{\theta^{(j)}} \mathcal{L}_{\mathrm{total}}$
\EndFor
\State \textbf{return} $F_{\tilde{\theta}}$ with $\tilde{\theta}=\theta^{(J)}$
\end{algorithmic}
\vspace{0.2em}
\hrule
\end{minipage}
\end{figure}

\subsubsection{Erasure Optimization Procedure} \label{sec:Erasure Optimization Procedure}

Let $\lambda_{\mathrm{TRA}}$ denote the weight of the task-to-rest response alignment loss. The total loss for erasure optimization is
\begin{equation}
\mathcal{L}_{\mathrm{total}} = \mathcal{L}_{\mathrm{pred}} + \lambda_{\mathrm{TRA}}\mathcal{L}_{\mathrm{TRA}}, 
\end{equation}
where $\mathcal L_{\mathrm{pred}}$ constrains model predictions and $\mathcal L_{\mathrm{TRA}}$ constrains selected Unit Neurons' responses.

During erasure optimization, we initialize $F_{\theta^{(0)}}$ from $F_{\theta}$ and minimize $\mathcal L_{\mathrm{total}}$ by updating the concept-responsive Unit Neurons' weights and biases using a binary parameter mask $M_{\mathcal{S}_c}$. All other parameters remain frozen. Let $\gamma$ denote the optimization step size, and let $\odot$ denote element-wise multiplication. An illustrative update is:
\begin{equation}
\theta^{(j+1)} = \theta^{(j)} - \gamma M_{\mathcal{S}_c} \odot \nabla_{\theta^{(j)}} \mathcal{L}_{\mathrm{total}}.
\end{equation} 

The complete CeFlune localization and erasure procedure is summarized in Algorithm~\ref{alg:CeFlune}.

\begin{table*}[!t]
\centering

\caption{Concept erasure performance on SDv1.4. CLIP is CLIPScore \cite{hessel2021clipscore} on the 200 MS-COCO prompts, used here to assess general image generation capability. \textbf{Bold} and \underline{underlined} values mark the highest and second-highest results in each column, respectively. \bad{Red} highlights pronounced degradation in mean CLIP.}
\label{tab:effectiveness_erasure}

\small
\setlength{\tabcolsep}{3.2pt}
\renewcommand{\arraystretch}{1.08}

\begin{tabular}{l cc cc cc cc cc cc | cc}

\toprule

\multirow{2}{*}{\textbf{Methods}}
& \multicolumn{2}{c}{\textbf{Nudity}}
& \multicolumn{2}{c}{\textbf{Van Gogh}}
& \multicolumn{2}{c}{\textbf{Church}}
& \multicolumn{2}{c}{\textbf{\makecell{Garbage\\Truck}}}
& \multicolumn{2}{c}{\textbf{Parachute}}
& \multicolumn{2}{c}{\textbf{Tench}}
& \multicolumn{2}{|c}{\textbf{Mean}}
\\

& ER$\uparrow$ & CLIP$\uparrow$
& ER$\uparrow$ & CLIP$\uparrow$
& ER$\uparrow$ & CLIP$\uparrow$
& ER$\uparrow$ & CLIP$\uparrow$
& ER$\uparrow$ & CLIP$\uparrow$
& ER$\uparrow$ & CLIP$\uparrow$
& ER$\uparrow$ & CLIP$\uparrow$
\\

\midrule

SDv1.4
& 25.0 & 25.75
& 0.0 & 25.75
& 1.5 & 25.75
& 3.0 & 25.75
& 23.5 & 25.75
& 32.0 & 25.75
& 14.17 & 25.75 \\

\midrule

FMN \cite{FMN}
& 26.0 & 25.61
& 62.5 & 25.80
& 31.0 & 25.81
& 30.0 & 25.78
& 47.5 & 25.46
& 70.5 & 25.56
& 44.58 & 25.67 \\

SPM \cite{SPM}
& 47.0 & \second{25.85}
& 53.5 & 25.70
& 23.0 & 25.82
& 81.0 & 25.74
& 73.0 & 25.77
& 83.0 & \second{25.85}
& 60.08 & \second{25.79} \\

UCE \cite{UCE}
& 79.0 & 25.80
& 43.5 & \best{26.22}
& 43.5 & 25.66
& 82.0 & \best{25.98}
& 70.5 & \second{25.91}
& 93.5 & 24.31
& 68.67 & 25.65 \\

AEGIS \cite{AEGIS}
& 57.5 & 25.22
& 83.0 & 24.26
& 77.5 & 24.36
& 98.0 & 23.90
& 89.0 & 24.33
& 44.0 & 24.31
& 74.83 & 24.40 \\

\midrule

CPE \cite{CPE}
& 82.0 & 25.84
& 13.0 & 25.77
& \second{99.5} & 25.83
& \best{100.0} & \second{25.81}
& \best{100.0} & 25.85
& \best{100.0} & 25.84
& 82.42 & \best{25.82} \\

ESD \cite{ESD}
& 68.5 & 25.02
& 90.0 & 25.20
& 73.0 & 25.21
& \best{100.0} & 24.36
& 90.0 & 25.09
& 94.5 & 25.05
& 86.00 & 24.99 \\

AGE \cite{AGE}
& \second{96.0} & 24.20
& 97.0 & 25.58
& 58.5 & 25.74
& 74.0 & 25.71
& 93.0 & 25.60
& \best{100.0} & 25.83
& 86.42 & 25.44 \\

SalUn \cite{SalUn}
& 90.5 & 23.05
& 75.5 & 25.83
& 85.0 & 25.99
& 87.5 & \second{25.81}
& 82.0 & \best{26.16}
& \second{99.5} & \best{26.13}
& 86.67 & 25.50 \\

ConceptPrune \cite{ConceptPrune}
& 85.5 & \best{25.94}
& 74.5 & \second{26.01}
& 76.0 & 25.24
& \second{99.5} & 25.20
& 93.5 & 25.34
& \best{100.0} & 25.63
& 88.17 & 25.56 \\

SLD-strong \cite{SLD}
& 43.5 & 24.33
& 94.0 & 24.47
& 96.0 & 25.16
& \second{99.5} & 24.73
& 98.5 & 25.09
& \best{100.0} & 25.04
& 88.58 & 24.80 \\

\midrule

SLD-max \cite{SLD}
& 58.0 & 23.80
& 97.0 & 23.70
& 96.0 & 24.77
& \best{100.0} & 24.08
& \second{99.5} & 24.87
& \best{100.0} & 25.17
& 91.75 & 24.40 \\

RECE \cite{rece}
& 82.0 & 25.64
& 83.0 & 25.98
& 90.0 & \second{26.00}
& \second{99.5} & 25.30
& 97.5 & 25.54
& \best{100.0} & 17.59
& 92.00 & 24.34 \\

AdvUnlearn \cite{AdvUnlearn}
& 83.0 & 23.65
& 92.0 & 25.75
& 92.5 & 25.49
& 96.5 & 25.56
& 98.5 & 25.59
& 98.5 & 25.45
& 93.50 & 25.25 \\

STEREO \cite{stereo}
& 87.5 & 24.86
& 94.5 & 25.25
& 97.5 & 24.41
& \best{100.0} & 24.47
& 98.5 & 24.17
& \best{100.0} & 25.28
& 96.33 & 24.74 \\

SAeUron \cite{SAeUron}
& \best{100.0} & 14.45
& \best{100.0} & 12.83
& \best{100.0} & 14.23
& \best{100.0} & 14.27
& \best{100.0} & 14.32
& \best{100.0} & 14.32
& \best{100.00} & \bad{14.07} \\

\midrule

\rowcolor{oursbg}
\textbf{CeFlune (ours)}
& 94.0 & 25.83
& \second{99.0} & 25.40
& 88.0 & \best{26.19}
& \best{100.0} & 25.28
& \best{100.0} & 19.58
& \best{100.0} & 25.14
& \second{96.83} & 24.57 \\

\bottomrule

\end{tabular}
\end{table*}

\begin{table*}[!t]
\centering

\caption{Concept erasure performance on SDv3. CLIP is CLIPScore \cite{hessel2021clipscore} on the 200 MS-COCO prompts, used here to assess general image generation capability. \textbf{Bold} and \underline{underlined} values mark the highest and second-highest results in each column, respectively. \bad{Red} highlights pronounced degradation in mean CLIP.}
\label{tab:effectiveness_erasure_sdv3}

\small
\setlength{\tabcolsep}{3.2pt}
\renewcommand{\arraystretch}{1.08}

\begin{tabular}{l cc cc cc cc cc cc | cc}

\toprule

\multirow{2}{*}{\textbf{Methods}}
& \multicolumn{2}{c}{\textbf{Nudity}}
& \multicolumn{2}{c}{\textbf{Van Gogh}}
& \multicolumn{2}{c}{\textbf{Church}}
& \multicolumn{2}{c}{\textbf{\makecell{Garbage\\Truck}}}
& \multicolumn{2}{c}{\textbf{Parachute}}
& \multicolumn{2}{c}{\textbf{Tench}}
& \multicolumn{2}{|c}{\textbf{Mean}}
\\

& ER$\uparrow$ & CLIP$\uparrow$
& ER$\uparrow$ & CLIP$\uparrow$
& ER$\uparrow$ & CLIP$\uparrow$
& ER$\uparrow$ & CLIP$\uparrow$
& ER$\uparrow$ & CLIP$\uparrow$
& ER$\uparrow$ & CLIP$\uparrow$
& ER$\uparrow$ & CLIP$\uparrow$
\\

\midrule

SDv3
& 52.5 & 25.37
& 32.5 & 25.37
& 33.5 & 25.37
& 10.0 & 25.37
& 12.5 & 25.37
& 47.5 & 25.37
& 31.42 & 25.37 \\

\midrule

RAZOR \cite{RAZOR}
& 48.5 & 25.30
& 30.5 & 25.32
& 32.0 & 25.33
& 7.5 & 25.42
& 18.5 & 25.40
& 67.5 & 25.43
& 34.08 & 25.37 \\

UCE \cite{UCE}
& 49.5 & \second{25.42}
& 38.5 & \second{25.45}
& 35.5 & 25.25
& 10.0 & 25.34
& 25.5 & 25.30
& 51.5 & 25.40
& 35.08 & 25.36 \\

HiRM \cite{HiRM}
& 50.5 & 25.22
& 31.0 & 25.30
& 36.5 & 25.40
& 11.5 & 25.32
& 19.0 & 25.41
& 62.5 & 25.31
& 35.17 & 25.33 \\

\midrule

CASteer \cite{casteer}
& \second{98.5} & 25.28
& 80.0 & 25.26
& 45.0 & \best{25.76}
& 7.0 & \second{25.48}
& 33.5 & \second{25.82}
& 76.0 & 25.46
& 56.67 & \second{25.51} \\

EraseAnything \cite{eraseanything}
& 82.5 & \best{25.46}
& 98.0 & 24.72
& 88.5 & 23.93
& 25.0 & 25.16
& 37.5 & 25.34
& 37.5 & \second{25.84}
& 61.50 & 25.08 \\

EraseAnything++ \cite{eraseanything++}
& 91.0 & 23.53
& 46.5 & 24.89
& 95.5 & 23.79
& 29.0 & 24.89
& 50.0 & 24.99
& 63.5 & 24.74
& 62.58 & 24.47 \\

SLD-strong \cite{SLD}
& 59.5 & 25.09
& 80.5 & 25.13
& 70.5 & 25.35
& 46.5 & 25.39
& 45.5 & 25.29
& 97.5 & 25.08
& 66.67 & 25.22 \\

\midrule

SalUn \cite{SalUn}
& 80.5 & 25.29
& 77.0 & \best{25.56}
& \second{99.0} & \second{25.73}
& \best{100.0} & \best{26.27}
& \best{100.0} & \best{26.07}
& \best{100.0} & \best{25.99}
& 92.75 & \best{25.82} \\

SLD-max \cite{SLD}
& 73.0 & 24.61
& 93.0 & 25.03
& 96.0 & 24.93
& \best{100.0} & 24.83
& 98.5 & 24.53
& \best{100.0} & 24.58
& 93.42 & 24.75 \\

ESD \cite{ESD}
& 85.0 & 25.37
& \best{100.0} & 14.54
& 92.5 & 24.42
& \second{99.5} & 21.66
& \second{99.5} & 17.88
& \second{98.5} & 24.29
& 95.83 & 21.36 \\

ConceptPrune \cite{ConceptPrune}
& \best{100.0} & 15.34
& 86.5 & 15.84
& \best{100.0} & 14.97
& \best{100.0} & 14.82
& \best{100.0} & 14.23
& \best{100.0} & 15.75
& \best{97.75} & \bad{15.16} \\

\midrule

\rowcolor{oursbg}
\textbf{CeFlune (ours)}
& 91.0 & 24.88
& \second{99.0} & 24.84
& 92.5 & 24.93
& \second{99.5} & 24.56
& 99.0 & 24.55
& \best{100.0} & 25.21
& \second{96.83} & 24.83 \\

\bottomrule

\end{tabular}

\end{table*}
\section{Experiments} \label{sec:Experiments}
We first assess CeFlune's ability to erase target concepts, its effects on other concepts, and the scope and cost of intervention. We then evaluate the ability of IVO and four other attacks to reactivate concepts erased by CeFlune and existing erasure methods, comparing the attacks' effectiveness and the erasure methods' resistance. We also observe if an erased concept can be deliberately restored in different erased models through subsequent fine-tuning. Finally, we present detailed ablation analyses for IVO and CeFlune.
\subsection{Experimental Setup}

\subsubsection{Models, Concepts, and Datasets} \label{sec:Models, Concepts, and Datasets}
Our evaluation spans different generative backbones and concept types. It uses screened prompt datasets to assess the erasure of target concepts and the preservation of other concepts, and a separate set of MS-COCO captions \cite{lin2014microsoft} to assess general image generation capability.

\textbf{Models.} Stable Diffusion 1.4\footnote{CompVis, \href{https://huggingface.co/CompVis/stable-diffusion-v1-4}{Stable Diffusion v1-4 Model Card}.} (SDv1.4) and Stable Diffusion 3 Medium\footnote{Stability AI, \href{https://huggingface.co/stabilityai/stable-diffusion-3-medium-diffusers}{Stable Diffusion 3 Medium Model Card}.} (SDv3) served as the two main evaluation models. SDv1.4 is a U-Net latent diffusion model, whereas SDv3 is a flow-matching model that uses a Multimodal Diffusion Transformer to predict a velocity field. For cross-architecture evaluation, we additionally included Stable Diffusion 2\footnote{Stability AI, \href{https://stability.ai/news-updates/stable-diffusion-v2-release}{Stable Diffusion 2.0 Release}.} (SDv2) and Stable Diffusion XL 1.0\footnote{Stability AI, \href{https://huggingface.co/stabilityai/stable-diffusion-xl-base-1.0}{SD-XL 1.0-base Model Card}.} (SDXL) as U-Net latent diffusion models, and FLUX.1-dev\footnote{Black Forest Labs, \href{https://huggingface.co/black-forest-labs/FLUX.1-dev}{FLUX.1-dev Model Card}.} (FLUX) as another flow-matching transformer.

\textbf{Concepts.} Following prior studies on concept erasure \cite{ESD,ConceptPrune}, we evaluated six core concepts: \textit{nudity}, \textit{Van Gogh style}, \textit{church}, \textit{garbage truck}, \textit{parachute}, and \textit{tench}. These concepts cover a broad range of erasure scenarios, including \textit{NSFW content}, \textit{artistic style}, and \textit{objects}. We also evaluated five other artist styles (\textit{Monet}, \textit{Picasso}, \textit{Dalí}, \textit{Munch}, and \textit{Warhol}) and five object concepts (\textit{apple}, \textit{cup}, \textit{sofa}, \textit{bicycle}, and \textit{dog}) to measure how erasure affects the generation of content that should be retained.

\textbf{Datasets.} For each concept, we created a lot of prompts with ChatGPT.\footnote{OpenAI, \url{https://chatgpt.com}.} We screened them by generating five images per prompt with the original SDv1.4 model, using a different seed for each image. A prompt was eligible only if the corresponding concept was detected in all five images. We randomly selected 200 eligible prompts per concept for evaluation. For assessment of general image generation capability after concept erasure, we randomly selected 200 distinct image IDs from the MS-COCO 2014 validation set and retained one caption per ID.

\begin{figure*}[!t]
    \centering
    \includegraphics[width=\textwidth]{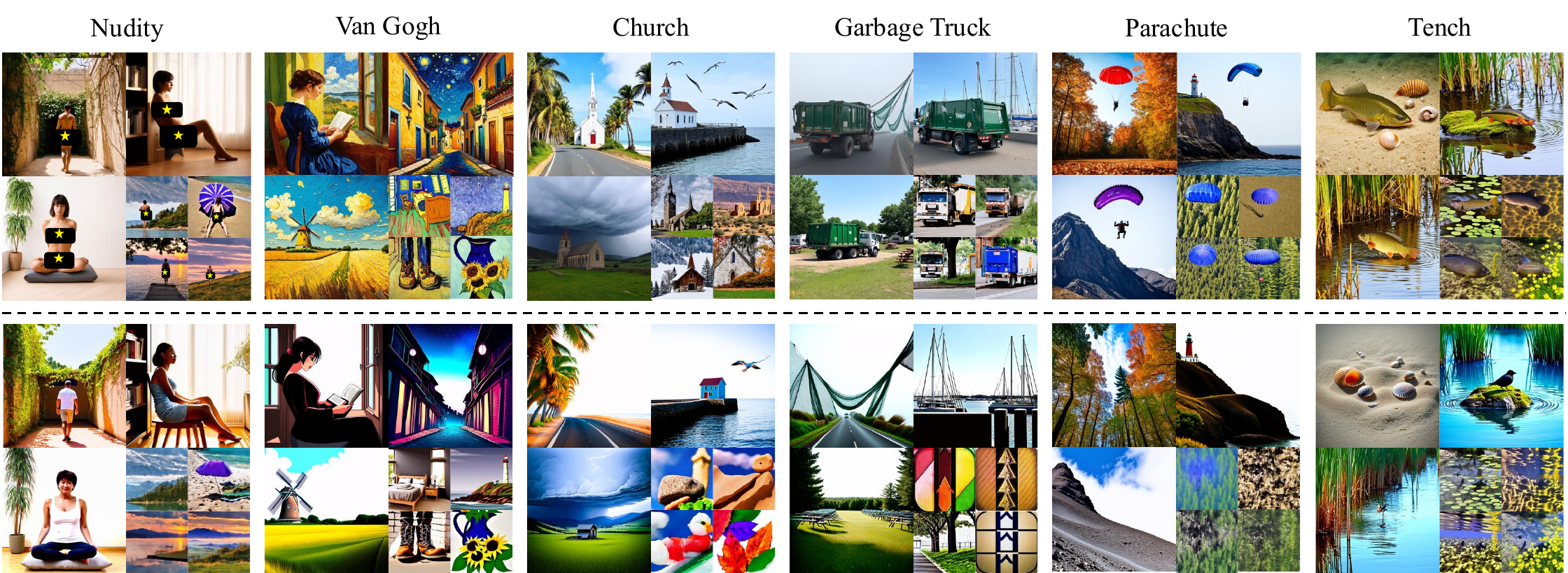}
    \caption{Erasure examples on SDv3 (large panels) and SDv1.4 (small panels). Images generated by the original model are shown above the dashed line, and those generated by CeFlune are shown below. \protect\colorbox{black}{\textcolor{yellow}{\ensuremath{\bigstar}}} indicates nudity masked for display.}
    \label{fig:erasure-examples}
\end{figure*}

\begin{figure}[t]
    \centering
    \includegraphics[width=\columnwidth]{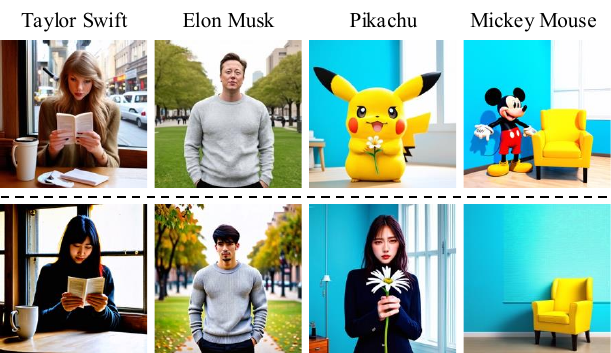}
    \caption{Erasure examples for celebrity identities and copyrighted characters on SDv3. Images generated by the original model are shown above the dashed line, and those generated by CeFlune are shown below.}
    \label{fig:celebrity_copyright_erasure}
\end{figure}

\subsubsection{Concept Detection and Evaluation Metrics} \label{sec:Concept Detection and Evaluation Metrics}
We use multiple detection methods and common metrics to assess concept erasure, preservation, and reactivation.

\textbf{Concept detection.} Following UnlearnDiffAtk \cite{UnlearnDiffAtk}, we used NudeNet\footnote{notAI-tech, \url{https://github.com/notAI-tech/NudeNet}.} to detect nudity. A detection score above 0.45 for any exposed body part \footnote{Female breast, female genitalia, male breast, male genitalia, buttocks, anus, feet, belly, and armpits.} indicated nudity. The artist classifier \cite{UnlearnDiffAtk} detected an artist style when its label appeared among the top 10 predictions. An ImageNet ResNet-50 classifier \cite{he2016deep} detected \textit{church}, \textit{garbage truck}, \textit{parachute}, and \textit{tench} when the corresponding class was its top prediction. Qwen2.5-VL \cite{bai2025qwen25vltechnicalreport} used a yes/no question to determine if the five other object concepts appeared in an image.

\textbf{Metrics.} We used the metrics defined in Section~\ref{sec:Preliminaries and Problem Definition}. Erasure rate (ER) measures the fraction of generated images in which the evaluated concept is absent. For the concept being erased, higher ER indicates stronger erasure. For concepts meant to remain, lower ER indicates better preservation. Attack success rate (ASR) measures how often the erased concept reappears under attack. Lower ASR indicates greater attack resistance. We also reported CLIPScore (CLIP) \cite{hessel2021clipscore}, which measures text–image semantic alignment \footnote{We use CLIP ViT-L/14 and report 100 times the cosine similarity between L2-normalized text and image embeddings.}. Higher CLIP scores indicate better text–image alignment.

\subsubsection{Compared Methods} \label{sec:Compared Methods}

\hspace*{\parindent}\textbf{Erasure baselines.} We compared CeFlune with 15 erasure baselines on SDv1.4 and 11 on SDv3. The SDv1.4 erasure baselines were ESD \cite{ESD}, FMN \cite{FMN}, SPM \cite{SPM}, CPE \cite{CPE}, AGE \cite{AGE}, AdvUnlearn \cite{AdvUnlearn}, STEREO \cite{stereo}, AEGIS \cite{AEGIS}, UCE \cite{UCE}, RECE \cite{rece}, SLD-strong, SLD-max \cite{SLD}, SalUn \cite{SalUn}, ConceptPrune \cite{ConceptPrune}, and SAeUron \cite{SAeUron}. The SDv3 erasure baselines were CASteer \cite{casteer}, SLD-strong, SLD-max, RAZOR \cite{RAZOR}, EraseAnything \cite{eraseanything}, EraseAnything++ \cite{eraseanything++}, HiRM \cite{HiRM}, ESD, SalUn, ConceptPrune, UCE.

For SDv1.4 baselines, we used official weights where available and otherwise reproduced the methods using the authors' released implementations and settings. For SDv3 baselines, we reproduced EraseAnything, EraseAnything++, HiRM, CASteer, and RAZOR using the authors' released code, and ESD, UCE, SalUn, and ConceptPrune by adapting their algorithms to SDv3.

\textbf{Attack baselines.} We compared IVO with Ring-A-Bell \cite{Ring-A-Bell}, P4D \cite{P4D}, UnlearnDiffAtk \cite{UnlearnDiffAtk}, and RECALL \cite{RECALL}. For each concept, we used the attack method's public configuration when available and its default settings otherwise.

\subsubsection{Implementation Details} \label{sec:IVO and CeFlune Implementation}

\hspace*{\parindent}\textbf{IVO implementation.} For each target prompt, we inverted five different reference images previously collected from the Internet to obtain the initial latents. We optimized each latent using the losses in Eq.~\ref{eq:ivo-objective}, with $t$ fixed at the 60\% position in the generation sequence and five updates per latent. We generated an image from each initial latent and after each update, producing 30 candidates per prompt. We then selected the image with the highest CLIPScore among those detected as containing the erased concept, or among all 30 candidates if none were detected.

\textbf{CeFlune implementation.} For each concept, we used the same text inputs across models to induce the task state (e.g., ``parachute'') and the resting state (empty text). To reduce noise in Unit Neuron response measurements, we separately averaged the task-state and resting-state response trajectories over the same 60 random seeds. We set $\epsilon=10^{-8}$ for the localization statistic $\psi$ and used a Spearman rank correlation test for $\mathcal T$. Unit Neurons were selected when the correlation was not statistically significant ($p\geq\tau$, $\tau=0.05$). If fewer than two were selected, we also included those with p-values below $\tau$ and negative correlation. We used Adam with a constant learning rate of $5\times10^{-5}$, batch size 1, and 600 updates. We set $\lambda_{\perp}=3$, $\lambda_{\mathrm{CDR}}=1$, and $\lambda_{\mathrm{TRA}}=0.1$. The negative-guidance strength $\eta$ was set to each model’s default guidance scale for image generation. At each erasure update, we randomly sampled one image from the neutral image collection of NSFW Data Scraper \footnote{alex000kim, \href{https://github.com/alex000kim/nsfw_data_scraper}{NSFW Data Scraper}.}.

Unless otherwise noted for an individual experiment, the settings in this section applied throughout. Concept erasure methods based on the same original generative model were run at the same numerical precision. All experiments were conducted on NVIDIA V100 GPUs with 32 GB of memory per GPU. Methods that could not fit on one GPU used multiple GPUs.

\subsection{Evaluation of Concept Erasure} \label{sec:Evaluation of Concept Erasure}

\subsubsection{Single-concept Erasure} \label{sec:Single-concept Erasure}
Table~\ref{tab:effectiveness_erasure} shows why concept erasure and general image generation capability must be assessed together. For SDv1.4, CeFlune raises mean ER from 14.17\% to 96.83\%, but only changes mean CLIP from 25.75 to 24.57. STEREO shows a similar balance, at 96.33\% mean ER and 24.74 mean CLIP. SAeUron reaches 100\% mean ER. However, its mean CLIP falls to 14.07, showing that higher ER can accompany a substantial decline in general image generation capability. 

Table~\ref{tab:effectiveness_erasure_sdv3} shows that on SDv3, CeFlune raises mean ER from 31.42\% to 96.83\%, while mean CLIP changes from 25.37 to 24.83. ConceptPrune has a higher mean ER than CeFlune (97.75\% vs. 96.83\%), but its mean CLIP is substantially lower by 9.67. ESD falls short of CeFlune on both metrics. SalUn surpasses CeFlune in mean CLIP with a score of 25.82, while trailing it in ER at 92.75\%. These comparisons support a favorable balance between concept erasure and general image generation capability for CeFlune.

Fig.~\ref{fig:erasure-examples} compares images generated by the original model and CeFlune for six core concepts on SDv1.4 and SDv3. Each comparison uses the same prompt and random seed. In these examples, the images generated after CeFlune erasure no longer show the target concept, although some other scene details differ from the images generated by the original model. Fig.~\ref{fig:celebrity_copyright_erasure} extends this visual comparison to celebrity identities and copyrighted characters.

\begin{figure}[h]
    \centering
    \includegraphics[width=\linewidth]{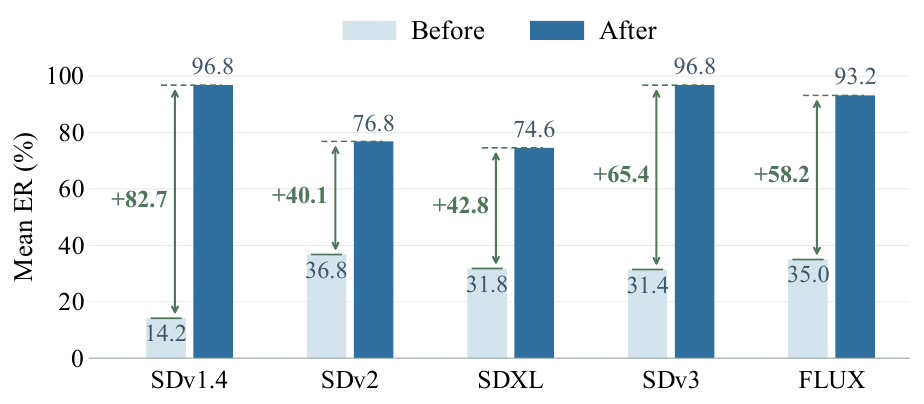}
    \caption{Mean erasure rate across the six core concepts on five model architectures before and after applying CeFlune.}
    \label{fig:cross_architecture}
\end{figure}

\subsubsection{Cross-architecture Evaluation} \label{sec:Cross-architecture Evaluation}
We applied CeFlune separately to SDv1.4, SDv2, SDXL, SDv3, and FLUX using the six core concepts. Fig.~\ref{fig:cross_architecture} shows that mean ER across the six concepts increases for all five models after applying CeFlune. The mean ER across the six concepts reaches 96.8\% on both SDv1.4 and SDv3, 93.2\% on FLUX, 76.8\% on SDv2, and 74.6\% on SDXL. These results support the applicability of CeFlune across the evaluated U-Net and transformer architectures.

\begin{table}[t]
\small
\centering
\caption{Erasure effectiveness and preservation of related and unrelated concepts. The Related and Unrelated columns report means over five other artist styles and five objects, respectively. \best{Bold} and \second{underlined} values mark the best and second-best results among erasure methods in each column. \bad{Red} highlights pronounced CLIP degradation.}
\label{tab:e8_preservation}
\begin{tabular}{lccccc}
\toprule
\multirow{2}{*}{\textbf{Methods}}
& \textbf{Target} 
& \multicolumn{2}{c}{\textbf{Related }} 
& \multicolumn{2}{c}{\textbf{Unrelated}} \\

\cmidrule(lr){2-2}
\cmidrule(lr){3-4}
\cmidrule(lr){5-6}

& \textbf{ER} $\uparrow$
& \textbf{ER} $\downarrow$ 
& \textbf{CLIP} $\uparrow$
& \textbf{ER} $\downarrow$ 
& \textbf{CLIP} $\uparrow$ \\
\midrule

SDv1.4 & 0.0 & 6.2 & 31.54 & 0.7 & 30.38 \\

\midrule

ESD & 90.0 & 37.0 & 29.05 & \best{0.7} & 28.98 \\

UCE & 43.5 & \best{12.8} & \best{31.10} & 1.1 & \best{30.58} \\

AdvUnlearn & 92.0 & 27.2 & \second{29.93} & \best{0.7} & \second{30.06} \\

ConceptPrune & 74.5 & 39.2 & 28.44 & \second{0.9} & 29.07 \\

SLD-max & \second{97.0} & 40.2 & 22.63 & 1.7 & 26.49 \\

SAeUron & \best{100.0} & 54.6 & 14.02 & 100.0 & \bad{14.87} \\

\midrule
\rowcolor{oursbg}
\textbf{CeFlune (ours)} & \best{100.0} & \second{22.8} & 24.93 & 3.0 & 27.38 \\

\bottomrule
\end{tabular}
\end{table}

\begin{table*}[!t]
\centering

\caption{Erasure effectiveness, intervention scope, and erasure cost across methods. The Localization column indicates whether a method includes an explicit concept localization step. For each method, the Neurons column counts the linear modules affected by its intervention, following CeFlune's Unit Neuron definition. The Params. column reports the number of parameters involved in the intervention. Neuron Ratio and Param. Ratio are calculated relative to the total numbers of Unit Neurons and parameters in the original SDv1.4 model, respectively, excluding parameters added by erasure methods. ER and ratios are percentages, parameter counts are in millions (M), and GPU memory is in GB. \best{Bold} and \second{underlined} values mark the best and second-best results in ER, Time, and GPU Mem.}
\label{tab:efficiency_comparison}

\small
\setlength{\tabcolsep}{3.3pt}
\renewcommand{\arraystretch}{1.10}

\begin{tabular}{l c c c c c c c c}

\toprule

\multirow[c]{2}{*}{\textbf{Method}} &
\multirow[c]{2}{*}{\textbf{Localization}} &
\multirow[c]{2}{*}{\textbf{ER $\uparrow$}} &
\multicolumn{4}{c}{\textbf{Intervention Scope}} &
\multicolumn{2}{c}{\textbf{Erasure Cost}} \\

\cmidrule(lr){4-7}
\cmidrule(lr){8-9}

& & & \textbf{Neurons} & \textbf{Neuron Ratio} & \textbf{Params.} & \textbf{Param. Ratio} & \textbf{Time} $\downarrow$ & \textbf{GPU Mem.} $\downarrow$\\

\midrule

\rowcolor{groupbg}
\multicolumn{9}{l}{\textit{Fine-tuning-based Methods}} \\

SalUn \cite{SalUn} & \ding{51} & 82.0 & 184 & 69.70 & 429.76 & 40.31 & \hrs{5.03} & 43.68 \\

ESD \cite{ESD} & \ding{55} & 90.0 & 120 & 45.45 & 815.56 & 76.49 & \hrs{3.45} & 15.74 \\

FMN \cite{FMN} & \ding{55} & 47.5 & 64 & 24.24 & 43.96 & 4.12 & \second{\secs{53}} & 12.37 \\

AdvUnlearn \cite{AdvUnlearn} & \ding{55} & \second{98.5} & 6 & 2.27 & 7.09 & 0.66 & \hrs{53.27} & 11.46 \\

STEREO \cite{stereo} & \ding{55} & \second{98.5} & 120 & 45.45 & 815.56 & 76.49 & \hrs{2.71} & 18.21 \\

AGE \cite{AGE} & \ding{55} & 93.0 & 120 & 45.45 & 815.56 & 76.49 & \hrs{6.99} & 16.83 \\

AEGIS \cite{AEGIS} & \ding{55} & 89.0 & 184 & 69.70 & 859.52 & 80.61 & \hrs{3.99} & 28.43 \\

SPM \cite{SPM} & \ding{55} & 73.0 & 160 & 60.61 & 1.70 & 0.16 & \hrs{3.20} & 8.14 \\

CPE \cite{CPE} & \ding{55} & \best{100.0} & 32 & 12.12 & 0.08 & 0.01 & \mins{16.54} & 8.71 \\

\midrule
\rowcolor{groupbg}
\multicolumn{9}{l}{\textit{Closed-form-based Methods}} \\

UCE \cite{UCE} & \ding{55} & 70.5 & 32 & 12.12 & 19.17 & 1.80 & \best{\secs{1.40}} & \best{5.27} \\

RECE \cite{rece} & \ding{55} & 97.5 & 32 & 12.12 & 19.17 & 1.80 & \mins{33.99} & \second{5.82} \\

\midrule
\rowcolor{groupbg}
\multicolumn{9}{l}{\textit{Pruning-based Methods}} \\

ConceptPrune \cite{ConceptPrune} & \ding{51} & 93.5 & 16 & 6.06 & 0.99 & 0.09 & \mins{23.07} & 9.31 \\

SAeUron \cite{SAeUron} & \ding{51} & \best{100.0} & 10 & 3.79 & 52.45 & 4.92 & \mins{30.95} & 18.76 \\

\midrule

\rowcolor{oursbg}
\textbf{CeFlune (ours)} & \ding{51} & \best{100.0} & 6 & 2.27 & 11.17 & 1.05 & \mins{36.40} & 9.01 \\

\bottomrule

\end{tabular}
\end{table*}

\subsubsection{Multi-concept Erasure} \label{sec:Multi-concept Erasure}
We evaluated multi-concept erasure with CeFlune on SDv1.4 using five concepts: \textit{nudity}, \textit{Van Gogh style}, \textit{church}, \textit{garbage truck}, and \textit{parachute}. We first localized Unit Neurons for each concept on SDv1.4 and then combined the selected Unit Neurons for optimization. At each optimization step, we computed the CeFlune erasure loss for each concept and averaged the five losses, but restricted each concept's task-to-rest response alignment loss to its own concept-responsive Unit Neurons.

CeFlune achieves a mean ER of 83.9\% across these five concepts, compared with 96.2\% for the corresponding single-concept erasures in Table~\ref{tab:effectiveness_erasure}. The difference is driven mainly by nudity and church, whose ERs are 65.0\% and 67.0\% under multi-concept erasure. The other three concepts reach at least 89.5\%. These results show that CeFlune can erase several concepts together, although its mean ER is notably lower than for the single-concept erasures.

\subsubsection{Preservation of Related and Unrelated Concepts} \label{sec:Preservation of Related and Unrelated Cocepts}

We use Van Gogh style erasure on SDv1.4 as a test case, with five other artist styles as related concepts and five objects as unrelated concepts. For each retained concept, we randomly selected 100 evaluation prompts from the dataset described in Section~\ref{sec:Models, Concepts, and Datasets}. Table~\ref{tab:e8_preservation} reports the preservation results for related and unrelated concepts.

SAeUron achieves 100.0\% ER for Van Gogh style, but its mean ER for unrelated concepts also reaches 100.0\%. Concepts that should have remained are erased as well, explaining its very low CLIP of 14.87. CeFlune also achieves 100.0\% ER for Van Gogh style. Its mean ERs for related and unrelated concepts are 22.8\% and 3.0\%, with CLIP scores of 24.93 and 27.38, respectively. Related and unrelated concept generation remains largely intact, indicating that CeFlune's erasure is focused on the target concept.

\subsubsection{Intervention Scope and Erasure Cost} \label{sec:Intervention Scope and Erasure Cost}

Table~\ref{tab:efficiency_comparison} compares the intervention scope, erasure effectiveness, and cost of the evaluated methods for erasing the parachute concept from SDv1.4. For each baseline, we count the Unit Neurons affected by its intervention using the same Unit Neuron definition as CeFlune. These counts describe intervention scope. The Localization column records whether a method explicitly localizes the concept.

CeFlune localizes 6 concept-responsive Unit Neurons and updates the corresponding 11.17M parameters, reaching 100.0\% ER. ESD and STEREO each affect 120 Unit Neurons and 815.56M parameters, but attain lower ER with longer erasure times and higher GPU memory use. CPE also achieves 100.0\% ER, but intervenes on more Unit Neurons than CeFlune (32 vs. 6). SAeUron reaches 100.0\% ER with 10 Unit Neurons, while its intervention involves more parameters (52.45M vs. 11.17M) and uses more GPU memory (18.76 GB vs. 9.01 GB). These results suggest that concept erasure can be achieved by localizing and updating concept-responsive Unit Neurons, without broad intervention across the model and with relatively low erasure costs.

\begin{table*}[!t]
\centering

\caption{Concept reactivation under five attacks on SDv1.4. For each attack, ASR and CLIP are means across the six core concepts. CLIP is CLIPScore between the attack-generated images and the original prompts.}
\label{tab:attack_results}

\small
\setlength{\tabcolsep}{4pt}
\renewcommand{\arraystretch}{1.10}

\begin{tabular}{l cc cc cc cc cc | cc}

\toprule

\multirow{2}{*}{\textbf{Erasure Method}}
& \multicolumn{2}{c}{\textbf{Ring-A-Bell} \cite{Ring-A-Bell}}
& \multicolumn{2}{c}{\textbf{P4D} \cite{P4D}}
& \multicolumn{2}{c}{\textbf{UnlearnDiffAtk} \cite{UnlearnDiffAtk}}
& \multicolumn{2}{c}{\textbf{RECALL} \cite{RECALL}}
& \multicolumn{2}{c}{\textbf{IVO (ours)}}
& \multicolumn{2}{|c}{\textbf{Mean}}
\\

\cmidrule(lr){2-3}
\cmidrule(lr){4-5}
\cmidrule(lr){6-7}
\cmidrule(lr){8-9}
\cmidrule(lr){10-11}
\cmidrule(lr){12-13}

& ASR & CLIP
& ASR & CLIP
& ASR & CLIP
& ASR & CLIP
& ASR & CLIP
& ASR & CLIP
\\

\midrule

ESD \cite{ESD}
& 18.54 & 15.29
& 6.88 & 11.20
& 58.11 & 24.79
& 85.63 & 25.29
& 94.17 & 30.01
& 52.67 & 21.32
\\

AdvUnlearn \cite{AdvUnlearn}
& 1.67 & 10.49
& 3.33 & 10.56
& 25.65 & 19.60
& 62.08 & 23.12
& 57.50 & 27.29
& 30.05 & 18.21
\\

UCE \cite{UCE}
& 23.33 & 18.77
& 17.08 & 12.28
& 89.47 & 27.29
& 90.83 & 27.30
& 86.46 & 31.16
& 61.43 & 23.36
\\

ConceptPrune \cite{ConceptPrune}
& 18.13 & 17.82
& 6.46 & 11.32
& 61.33 & 24.95
& 73.54 & 26.55
& 68.96 & 30.37
& 45.68 & 22.20
\\

STEREO \cite{stereo}
& 2.92 & 11.53
& 2.92 & 8.00
& 24.84 & 19.89
& 50.21 & 20.95
& 59.58 & 26.78
& 28.09 & 17.43
\\

RECE \cite{rece}
& 4.79 & 15.11
& 11.46 & 13.04
& 55.93 & 23.75
& 66.46 & 24.82
& 61.25 & 28.67
& 39.98 & 21.08
\\

SalUn \cite{SalUn}
& 11.25 & 16.43
& 4.58 & 12.02
& 55.55 & 23.52
& 71.04 & 23.89
& 65.00 & 28.68
& 41.48 & 20.91
\\

AEGIS \cite{AEGIS}
& 19.38 & 14.81
& 9.79 & 11.94
& 72.50 & 24.40
& 96.67 & 25.31
& 97.29 & 30.29
& 59.13 & 21.35
\\

\midrule
\rowcolor{oursbg}
\textbf{CeFlune (ours)}
& 0.0 & 11.17
& 3.33 & 10.17
& 26.59 & 20.21
& 45.00 & 22.07
& 43.54 & 26.97
& 23.69 & 18.12
\\

\midrule
\textbf{Mean}
& 11.11 & 14.60
& 7.31 & 11.17
& 52.22 & 23.16 
& 71.27 & 24.37
& 70.42 & 28.91
& — & —
\\

\bottomrule

\end{tabular}
\end{table*}

\subsection{Evaluation of Concept Reactivation}  \label{sec:Evaluation of Concept Reactivation}
\subsubsection{Reactivation under Attacks} \label{sec:Reactivation under Attacks}
We erased each of the six core concepts from SDv1.4 with nine methods, then attacked the erased models with Ring-A-Bell \cite{Ring-A-Bell}, P4D \cite{P4D}, UnlearnDiffAtk \cite{UnlearnDiffAtk}, RECALL \cite{RECALL}, and IVO. For each concept, we randomly selected 80 of the 200 evaluation prompts in Section~\ref{sec:Models, Concepts, and Datasets} and used them throughout the attack evaluation. Table~\ref{tab:attack_results} reports ASR and CLIP for each combination of attack and erasure method.

IVO achieves ASR above 50\% against all eight baseline erasure methods, reaching 97.29\% against AEGIS. Across the nine erased models, IVO's mean ASR is 70.42\%, close to RECALL's 71.27\%. IVO also produces the highest CLIP among the five attacks for each erased model, with values from 26.78 to 31.16. These results show that IVO exposes a new vulnerability in concept erasure: concepts can be reactivated by optimizing the initial latent, even when the target prompt remains unchanged.

Across the five attacks, CeFlune has the lowest mean ASR (23.69\%), compared with 28.09\% for STEREO. It also achieves the lowest ASR under Ring-A-Bell (0\%), RECALL (45.00\%), and IVO (43.54\%). These results suggest that CeFlune makes erased concepts harder to reactivate across attacks conducted in different optimization spaces.

\begin{figure}[t]
    \centering
    \includegraphics[width=\columnwidth]{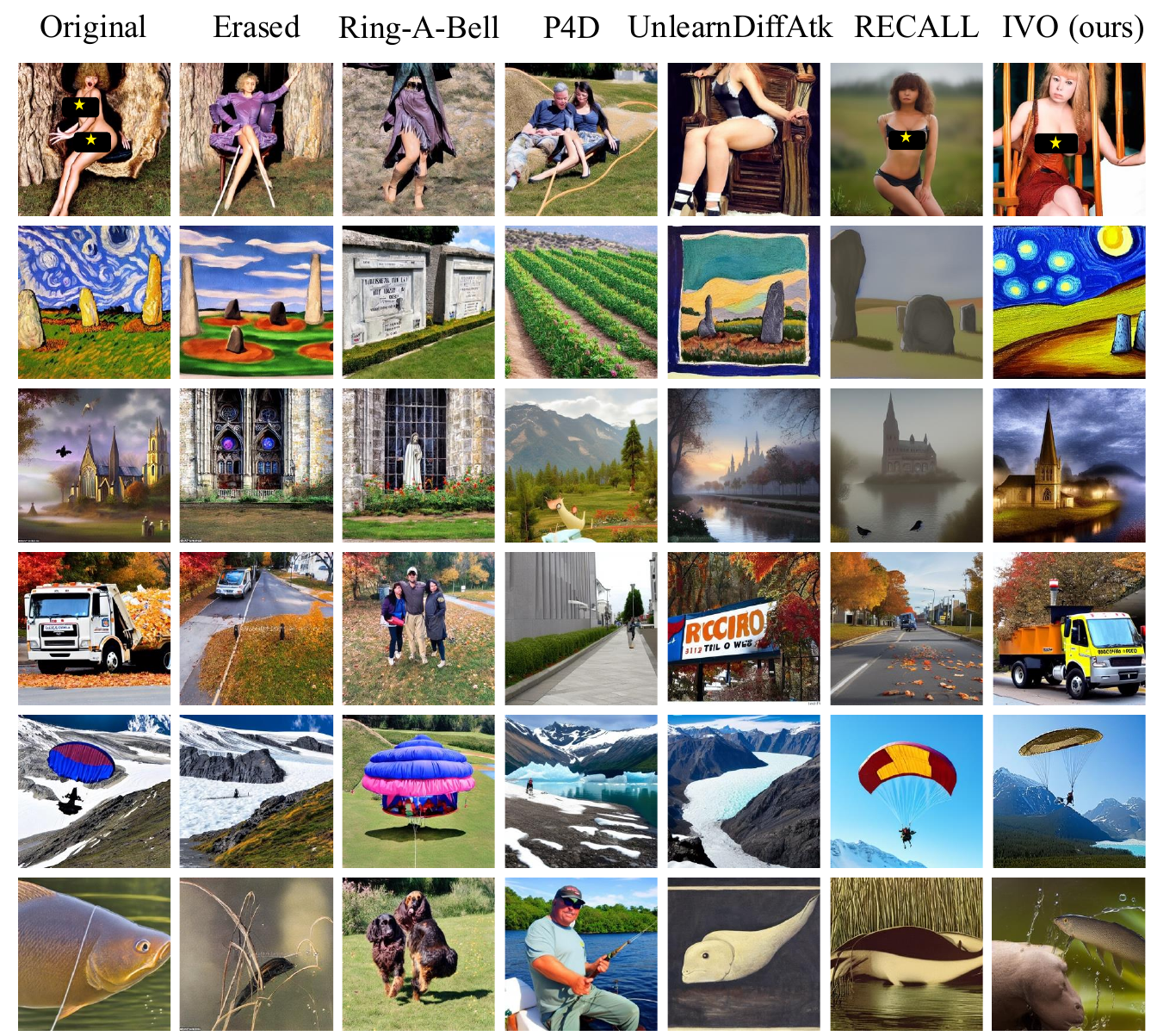}
    \caption{Visual comparison of concept reactivation attacks on SDv1.4 after ESD erasure. The Original and Erased columns show images generated without an attack by the original and erased models, respectively. \protect\colorbox{black}{\textcolor{yellow}{\ensuremath{\bigstar}}} indicates nudity masked for display.}
    \label{fig:attack_visualization}
\end{figure}

Fig.~\ref{fig:attack_visualization} compares the attack outputs with images from the original model. In several examples, IVO restores the erased concept while retaining scene details visible in the original images. This visual pattern is consistent with IVO's higher CLIP scores in Table~\ref{tab:attack_results}.

\begin{figure}[t]
    \centering
    \includegraphics[width=\linewidth]{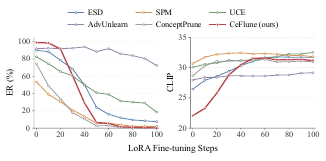}
    \caption{Concept reactivation during LoRA fine-tuning after erasure. Lower ER indicates more frequent detection of the erased concept.}
    \label{fig:lora_recovery_vangogh}
\end{figure}

\subsubsection{Reactivation under Fine-Tuning} \label{sec:Reactivation under Fine-Tuning}
We used subsequent LoRA training to deliberately restore Van Gogh style in SDv1.4 models erased by ESD, SPM, UCE, AdvUnlearn, ConceptPrune, and CeFlune. For each erased model, we added new LoRA modules to the query, key, value, and output projections of the U-Net attention layers, with rank 4, LoRA alpha 4, and dropout 0. We froze each erased model and trained the newly added LoRA parameters using the standard diffusion denoising loss. The training set comprised 2,000 Van Gogh image-caption pairs, generated by the original SDv1.4 model from 200 training prompts with 10 images per prompt. Each run used the same training set for 100 AdamW optimization steps (a constant learning rate of $10^{-4}$, and a batch size of 1).

Fig.~\ref{fig:lora_recovery_vangogh} shows Van Gogh ER from step 0 to step 100 on the Van Gogh evaluation prompts described in Section~\ref{sec:Models, Concepts, and Datasets}. CeFlune's Van Gogh ER falls from 99\% at step 0 to 6.5\% at step 50 and reaches 0\% at step 100. AdvUnlearn starts at 92\% ER and retains 72.5\% at step 100. At that point, ESD, SPM, UCE, and ConceptPrune have ER values between 1.5\% and 18.5\%. Van Gogh style can be rapidly restored in several erased models by training new LoRA parameters, while the erased model parameters remain fixed. This may be useful when the requirement to erase a concept changes.

\subsection{Ablation Analysis of IVO}\label{sec:Ablation Analysis of IVO}

\subsubsection{Initialization, Loss, and Time Ablations}
We ablate IVO on SDv1.4 models after concept erasure with ESD, using the attack prompts from Section~\ref{sec:Reactivation under Attacks}. Table~\ref{tab:ivo_ablation} reports the mean ASR and CLIP across the six core concepts.

Under the default setting, IVO achieves 95.42\% ASR and 29.99 CLIP. Replacing reference-based initialization with random initialization reduces ASR to 60.83\%, with CLIP of 29.73. Removing $\mathcal L_{\mathrm{react}}$ or $\mathcal L_{\mathrm{anchor}}$ reduces ASR to 90.42\% and 92.92\%, respectively, with corresponding CLIP scores of 29.83 and 29.97. These results show that reference-based initialization and both loss terms contribute to concept reactivation in this setting.

We vary the time $t$ used in the IVO losses across the 10\%, 30\%, 50\%, 70\%, and 90\% positions in the generation sequence. Mean ASR ranges from 90.21\% to 93.96\%, with CLIP of 29.50–30.07, but the default setting achieves a higher ASR of 95.42\% with CLIP of 29.99. These results show that the choice of $t$ only slightly affects concept reactivation.

\begin{table}[t]
\centering
\caption{Ablation of IVO on ESD-erased SDv1.4. Results are averaged over the six core concepts. CLIP is CLIPScore between the attack-generated images and the original prompts. In Ref., \ding{51} and \ding{55} denote reference-based and random initialization, respectively. Percentages in Time indicate the position of $t$ in the generation sequence.}
\label{tab:ivo_ablation}
\small
\begin{tabular}{cccccc}
\toprule
\textbf{Ref.} & $\mathcal L_{\mathrm{react}}$
& $\mathcal L_{\mathrm{anchor}}$ & \textbf{Time}
& \textbf{ASR} $\uparrow$ & \textbf{CLIP} $\uparrow$ \\
\midrule
\ding{55} & \ding{51} & \ding{51} & 60\% & 60.83 & 29.73 \\
\ding{51} & \ding{55} & \ding{51} & 60\% & 90.42 & 29.83 \\
\ding{51} & \ding{51} & \ding{55} & 60\% & 92.92 & 29.97 \\
\midrule
\ding{51} & \ding{51} & \ding{51} & 10\% & 90.83 & 29.65 \\
\ding{51} & \ding{51} & \ding{51} & 30\% & 93.75 & 29.76 \\
\ding{51} & \ding{51} & \ding{51} & 50\% & 93.96 & 30.07 \\
\ding{51} & \ding{51} & \ding{51} & 70\% & 90.21 & 29.92 \\
\ding{51} & \ding{51} & \ding{51} & 90\% & 91.04 & 29.50 \\
\midrule
\rowcolor{oursbg}
\ding{51} & \ding{51} & \ding{51} & 60\% & 95.42 & 29.99 \\
\bottomrule
\end{tabular}
\end{table}

\subsubsection{Similarity to Reference Images}
\begin{figure}[t]
    \centering
    \includegraphics[width=\columnwidth]{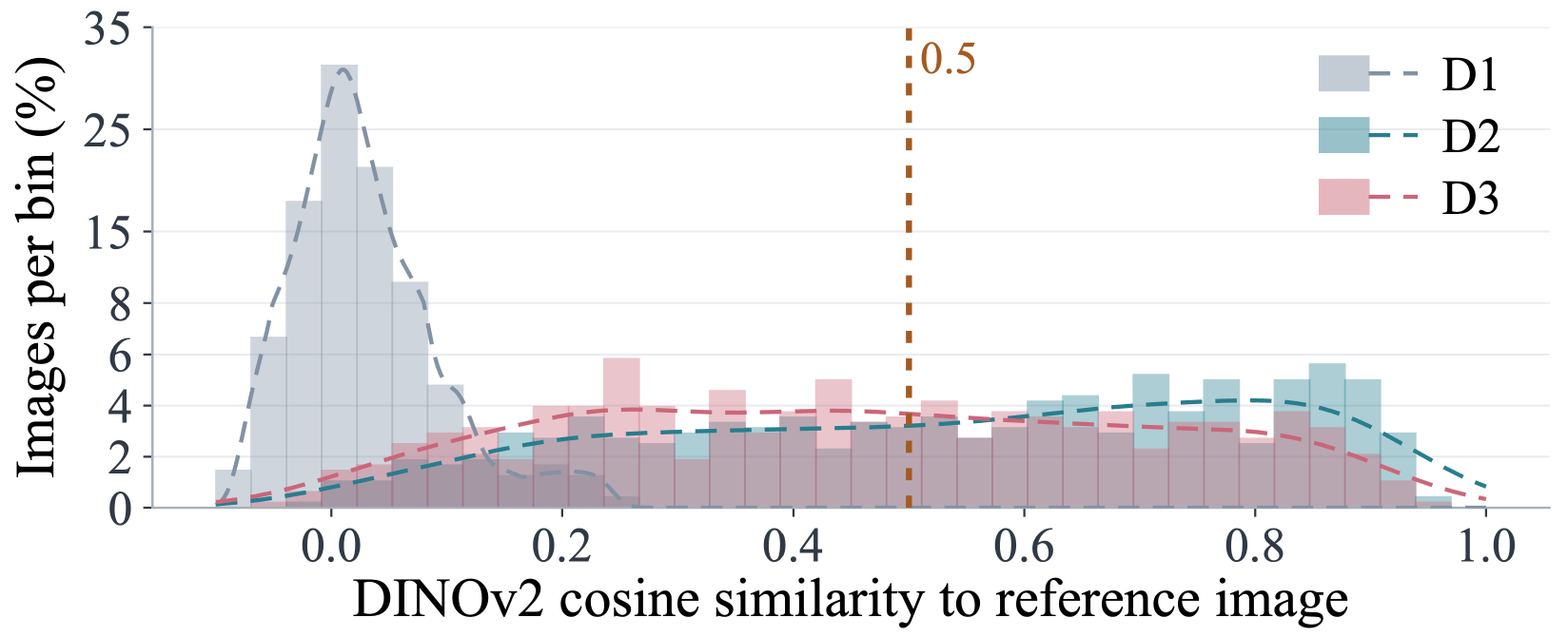}
    \caption{Similarity distributions between generated images and their paired reference images. The original SDv1.4 generates D1 images from random initial latents under an empty prompt and D2 images from reference-based initial latents under target prompts. D3 follows the D2 setting, except that IVO optimizes the initial latents and the erased model generates the images.}
    \label{fig:ivo_similarity_distribution}
\end{figure}

We extract image features with DINOv2 \cite{oquab2023dinov2} and compute the cosine similarity between each generated image and its paired reference image. Fig.~\ref{fig:ivo_similarity_distribution} compares the similarity distributions for D1, D2, and D3. Each distribution combines results from the six concepts. D1's narrow peak near zero provides a low-similarity baseline. D2 extends to higher similarities than D1, indicating greater resemblance to the paired reference images. D3 extends beyond D1's near-zero peak, contains fewer high-similarity images than D2, and does not cluster near 1. This pattern indicates that IVO outputs retain some visual resemblance to the reference images without directly reproducing them in full.

\subsection{Ablation Analysis of CeFlune} \label{sec:Ablation Analysis}
\subsubsection{Stability of Model Functional Localization} \label{sec:Stability of Model Functional Localization}

\begin{figure}[t]
    \centering
    \includegraphics[width=\linewidth]{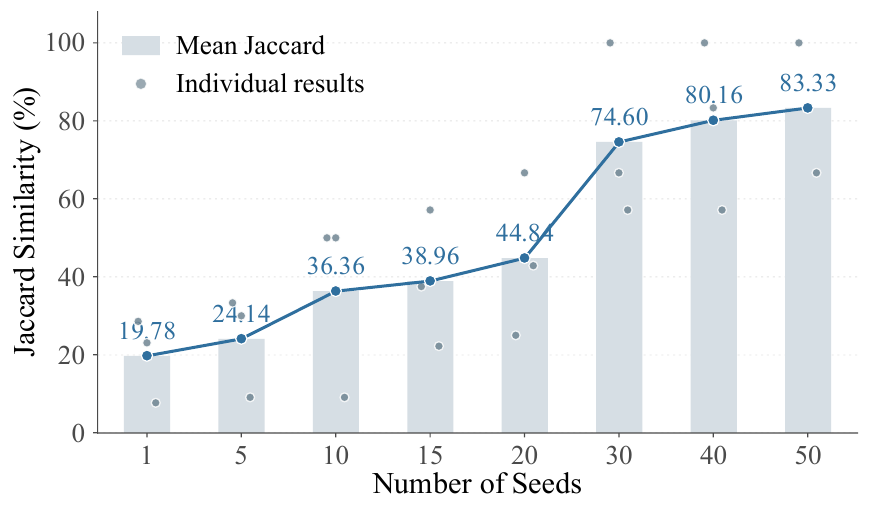}
    \caption{Stability of model functional localization as more seeds are used. Jaccard similarity is calculated between the concept-responsive Unit Neuron set selected using each seed subset and the set selected using all 60 seeds. Bars and the line show the mean across three seed orderings, and dots show the individual results.}
    \label{fig:selection_stability}
\end{figure}

We measured localization stability for the parachute concept on SDv1.4 using three fixed orderings of the same 60 seeds. Within each ordering, we increased the number of seeds used to average response trajectories, keeping other localization settings fixed. Fig.~\ref{fig:selection_stability} uses Jaccard similarity to compare the concept-responsive Unit Neurons selected using each seed subset with those selected using all 60 seeds.

Mean Jaccard similarity rises from 19.78\% with one seed to 44.84\% with 20 seeds, 74.60\% with 30 seeds, and 83.33\% with 50 seeds. At 50 seeds, Jaccard similarities to the 60-seed reference range from 66.67\% to 100\% across the three orderings. These results indicate that the concept-responsive Unit Neuron sets selected by CeFlune's model functional localization stabilize as more seeds are used.

\subsubsection{Cross-Concept Overlap of Functional Localization} \label{sec:Cross-concept Overlap}
\begin{figure}[t]
    \centering
    \includegraphics[width=\linewidth]{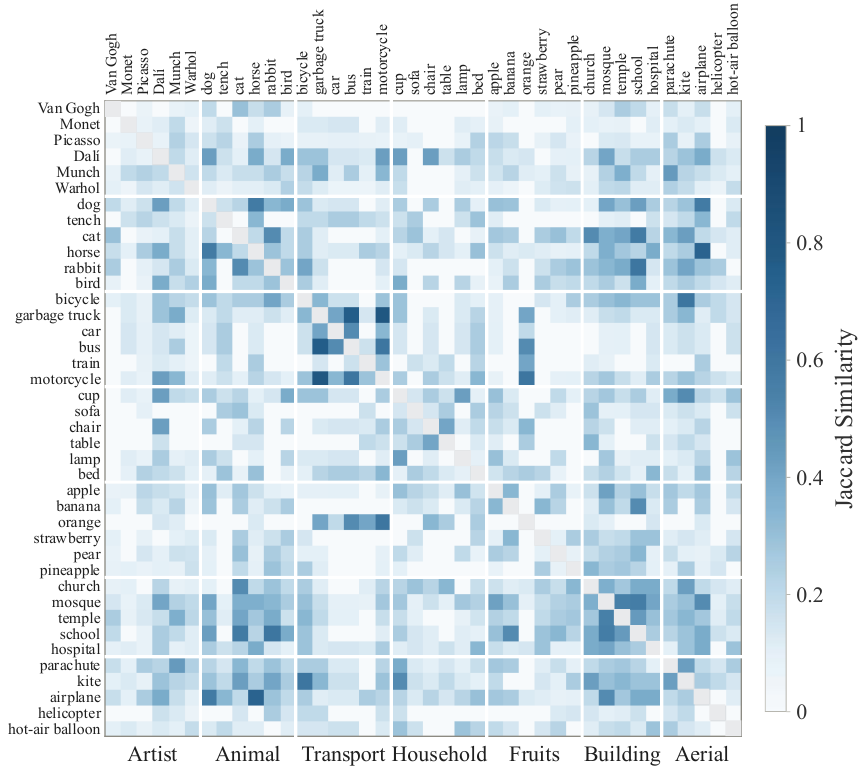}
    \caption{Jaccard similarity between the sets of concept-responsive Unit Neurons identified for 40 concepts in SDv1.4. Diagonal cells are shown in a neutral color.}
    \label{fig:cross_concept_overlap}
\end{figure}

We applied CeFlune's localization procedure to identify concept-responsive Unit Neurons for 40 concepts from seven categories in SDv1.4. Fig.~\ref{fig:cross_concept_overlap} compares these localization results using Jaccard similarity. The building concepts form a visible block of relatively high similarity, but the degree of overlap varies within the other categories. Some concepts from different categories also have concept-responsive Unit Neurons in common, with Jaccard similarity reaching 0.71 for certain pairs. These results suggest that indiscriminately pruning those Unit Neurons or zeroing their weights could affect more than one concept, making fine-tuning them a more suitable intervention.

\subsubsection{Visual Comparison of Functional Localization}
\begin{figure}[t]
    \centering
    \includegraphics[width=\columnwidth]{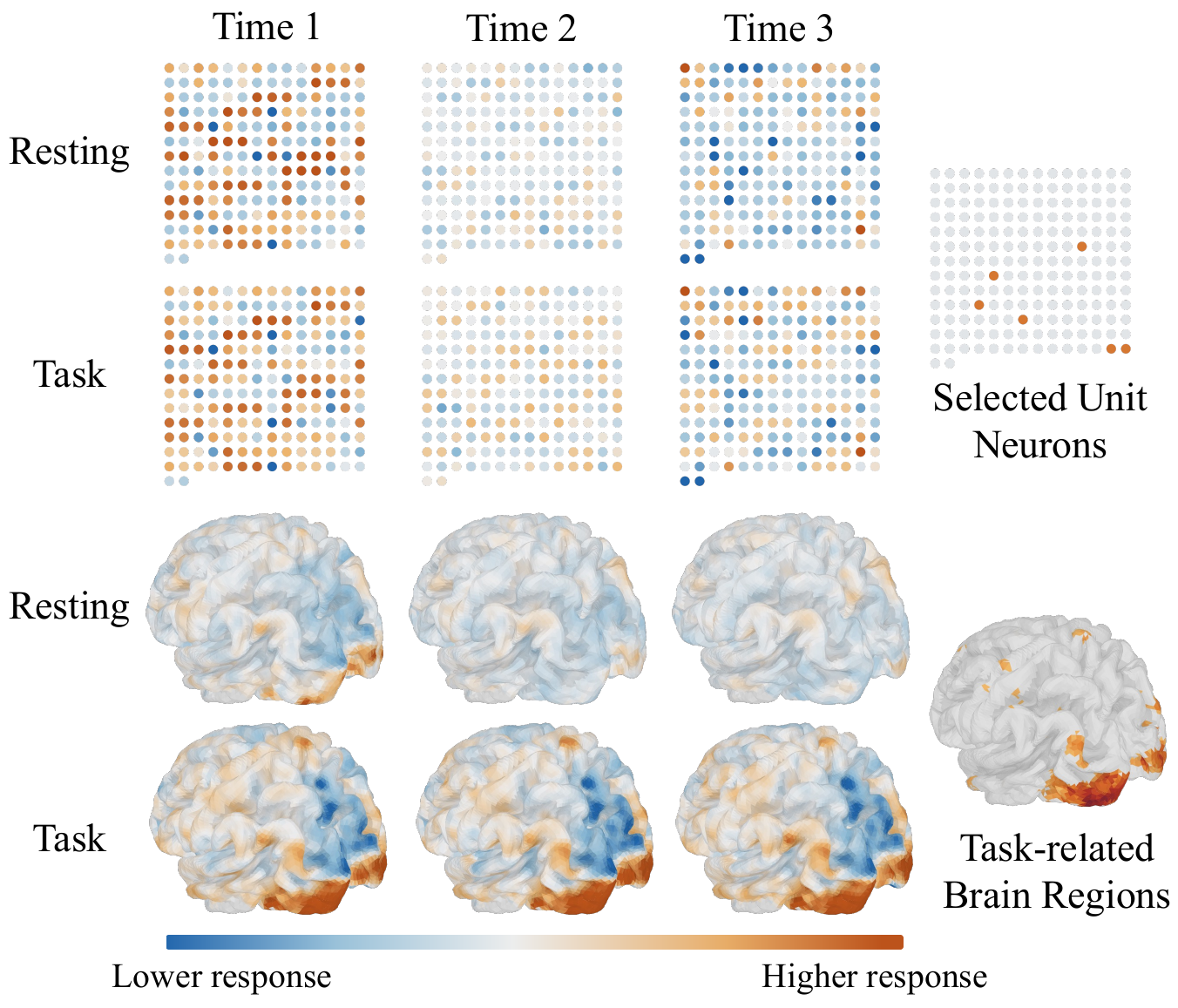}
    \caption{Visual comparison of model and brain functional localization for the church concept. The upper two rows show SDv1.4 Unit Neuron responses under empty text (Resting) and the ``church'' prompt (Task). The lower two rows show one CRSF \cite{wang2022fmri} participant’s fMRI activity during intervals without concept words or images (Resting) and after presentation of the church word and image (Task). CeFlune selects six concept-responsive Unit Neurons in SDv1.4, while fMRI identifies task-related brain regions in the participant.}
    \label{fig:church_functional_localization}
\end{figure}

Fig.~\ref{fig:church_functional_localization} visualizes functional localization for the church concept in SDv1.4 and human fMRI data, showing task-state and resting-state responses alongside the selected Unit Neurons and task-related brain regions. For the fMRI data, we use the concept representation with semantic features (CRSF) dataset \cite{wang2022fmri}, which recorded brain responses as participants viewed concept words paired with related images and thought about their meanings. The scans also included resting intervals during which participants viewed a screen without concept words or images. Time 1–3 show SDv1.4 Unit Neuron responses at generation steps 1, 25, and 50 under the “church” prompt and empty text, and fMRI activity at 4, 6, and 8 seconds after the church word and image are presented or a resting interval begins.

As shown in Fig.~\ref{fig:church_functional_localization}, the response patterns of candidate Unit Neurons change across Time 1–3. Responses at any one of these times alone cannot identify which Unit Neurons are concept-responsive. CeFlune compares each candidate Unit Neuron's task-state and resting-state response trajectories across the generation process, selecting six concept-responsive Unit Neurons for the church concept. For the participant shown in Fig.~\ref{fig:church_functional_localization}, brain activity is visible during resting intervals, but activity in regions toward the back of the brain appears more pronounced after the church word and image are presented. The fMRI localization compares brain activity during the church task with that during the resting intervals and identifies task-related brain regions in this participant. This visual comparison suggests that CeFlune can identify concept-responsive Unit Neurons amid widespread model responses, much as fMRI identifies task-related brain regions amid ongoing brain activity.

\subsubsection{Localization, Intervention and Loss Ablations}
Table~\ref{tab:ablation} shows how localization, intervention, and loss terms affect CeFlune’s results. Across six concepts, the full method averages 96.83\% ER, 24.57 CLIP, and 43.54\% ASR.

We replaced the Unit Neurons selected by CeFlune with random sets matched by neuron count or approximately by trainable parameter count. CeFlune achieves higher ER and lower ASR than both random baselines, which reach 67.11\% and 74.47\% ER and 88.61\% and 85.42\% ASR, respectively. Its CLIP score remains close to theirs (24.57 vs. 25.09 and 25.16). This comparison shows that accurate localization of concept-responsive Unit Neurons matters for erasure effectiveness and attack resistance.

We applied two alternative interventions to the Unit Neurons localized by CeFlune: zeroing their weights or replacing their task-state outputs with resting-state outputs from the original SDv1.4 model. CeFlune achieves higher ER and CLIP and lower ASR than both alternatives, which reach 87.50\% and 83.33\% ER, 21.47 and 17.45 CLIP, and 88.33\% and 83.96\% ASR, respectively. This comparison shows that how we intervene on concept-responsive Unit Neurons is also important.

We disable each of CeFlune's four loss terms in turn. Disabling $\mathcal L_{\perp}$, $\mathcal L_{\parallel}$, or $\mathcal L_{\mathrm{TRA}}$ lowers ER to 96.25\%, 89.42\%, and 93.25\%, respectively, and raises ASR to 51.67\%, 55.00\%, and 56.67\%. Disabling $\mathcal L_{\mathrm{CDR}}$ lowers ER most sharply, to 71.58\%, and raises ASR to 95.63\%. CeFlune's CLIP of 24.57 is close to the ablated variants' 24.75–25.80. Each loss term contributes to erasure and attack resistance, without substantially impairing general image generation capability.

\begin{table}[t]
\centering
\caption{Ablation of Unit Neuron selection, intervention, and loss terms. The Random (number matched) row uses the same number of Unit Neurons as CeFlune, and the Random (parameter matched) row uses approximately the same number of trainable parameters. Both rows report means across three random seeds. CLIP is CLIPScore on the 200 MS-COCO prompts, used to assess general image generation capability. ASR is measured under IVO.}
\label{tab:ablation}
\small
\begin{tabular}{lccc}
\toprule
\textbf{Variant}
& \textbf{ER} $\uparrow$
& \textbf{CLIP} $\uparrow$
& \textbf{ASR} $\downarrow$ \\
\midrule

\rowcolor{groupbg}
\multicolumn{4}{l}{\textit{Localization Ablation}} \\
Random (number matched) & 67.11 & 25.09 & 88.61 \\
Random (parameter matched) & 74.47 & 25.16 & 85.42 \\

\midrule
\rowcolor{groupbg}
\multicolumn{4}{l}{\textit{Intervention Ablation}} \\
Weight zeroing & 87.50 & 21.47 & 88.33 \\
Output replacement & 83.33 & 17.45 & 83.96 \\

\midrule
\rowcolor{groupbg}
\multicolumn{4}{l}{\textit{Objective Ablation}} \\
w/o $\mathcal{L}_{\perp}$ & 96.25 & 25.46 & 51.67 \\
w/o $\mathcal{L}_{\mathrm{CDR}}$ & 71.58 & 25.80 & 95.63 \\
w/o $\mathcal{L}_{\mathrm{TRA}}$ & 93.25 & 24.75 & 56.67 \\
w/o $\mathcal{L}_{\parallel}$ & 89.42 & 25.26 & 55.00 \\

\midrule
\rowcolor{oursbg}
\textbf{CeFlune} & 96.83 & 24.57 & 43.54 \\
\bottomrule
\end{tabular}
\end{table}

\begin{table}[t]
\centering
\caption{Sensitivity of CeFlune to three loss weights. Each weight is varied separately while the other two remain at their default values. The three 1$\times$ rows repeat the result with all weights at their default values. CLIP is CLIPScore on the 200 MS-COCO prompts, used to assess general image generation capability. ASR is measured under IVO.}
\label{tab:hyperparameter_sensitivity}
\small

\begin{tabular}{ccccc}
\toprule
\textbf{Parameter}
& \textbf{Scale}
& \textbf{ER} $\uparrow$
& \textbf{CLIP} $\uparrow$
& \textbf{ASR} $\downarrow$ \\
\midrule

\multirow{5}{*}{\makecell{$\lambda_{\perp}$\\{\scriptsize (default $=3$)}}}
& $0.25\times$ & 97.00 & 24.35 & 42.71 \\
& $0.5\times$  & 95.75  & 25.42 & 57.08 \\
& $1\times$    & 96.83 & 24.57 & 43.54 \\
& $2\times$    & 90.08 & 25.44 &  53.54 \\
& $4\times$    & 94.33  & 25.13 & 51.46 \\

\midrule

\multirow{5}{*}{\makecell{$\lambda_{\mathrm{CDR}}$\\{\scriptsize (default $=1$)}}}
& $0.25\times$ & 92.42  & 25.48 & 50.63 \\
& $0.5\times$  & 97.42 & 25.14 & 44.17 \\
& $1\times$    & 96.83 & 24.57 & 43.54 \\
& $2\times$    & 94.00 & 25.23 &  43.54 \\
& $4\times$    & 98.50 & 25.42 &  39.38 \\

\midrule

\multirow{5}{*}{\makecell{$\lambda_{\mathrm{TRA}}$\\{\scriptsize (default $=0.1$) }}}
& $0.25\times$ & 94.83 & 24.45 & 47.29 \\
& $0.5\times$  & 98.17 & 24.42 & 44.58 \\
& $1\times$    & 96.83 & 24.57 & 43.54 \\
& $2\times$    & 90.75 & 24.90 & 49.38 \\
& $4\times$    & 96.17 & 24.57 & 44.38 \\

\bottomrule
\end{tabular}
\end{table}

\subsubsection{Hyperparameter Sensitivity}

Table~\ref{tab:hyperparameter_sensitivity} shows the effects of the three loss weights on CeFlune. With the default setting, CeFlune achieves 96.83\% ER, 24.57 CLIP, and 43.54\% ASR across the six concepts. Reducing $\lambda_{\mathrm{CDR}}$ to 0.25× decreases ER to 92.42\% and increases ASR to 50.63\%, consistent with the loss ablation. At 4×, ER rises to 98.50\%, CLIP rises to 25.42, and ASR falls to 39.38\%. As $\lambda_{\mathrm{TRA}}$ varies from 0.25× to 4×, ER ranges from 90.75\% to 98.17\% and ASR from 43.54\% to 49.38\%. For $\lambda_{\perp}$, settings with CLIP above the default also have higher ASR. At 0.5×, ER is 95.75\%, CLIP rises to 25.42, and ASR to 57.08\%. At 0.25×, ER is 97.00\%, CLIP falls to 24.35, and ASR to 42.71\%.

\begin{figure}[!t]
    \centering
    \includegraphics[width=\linewidth]{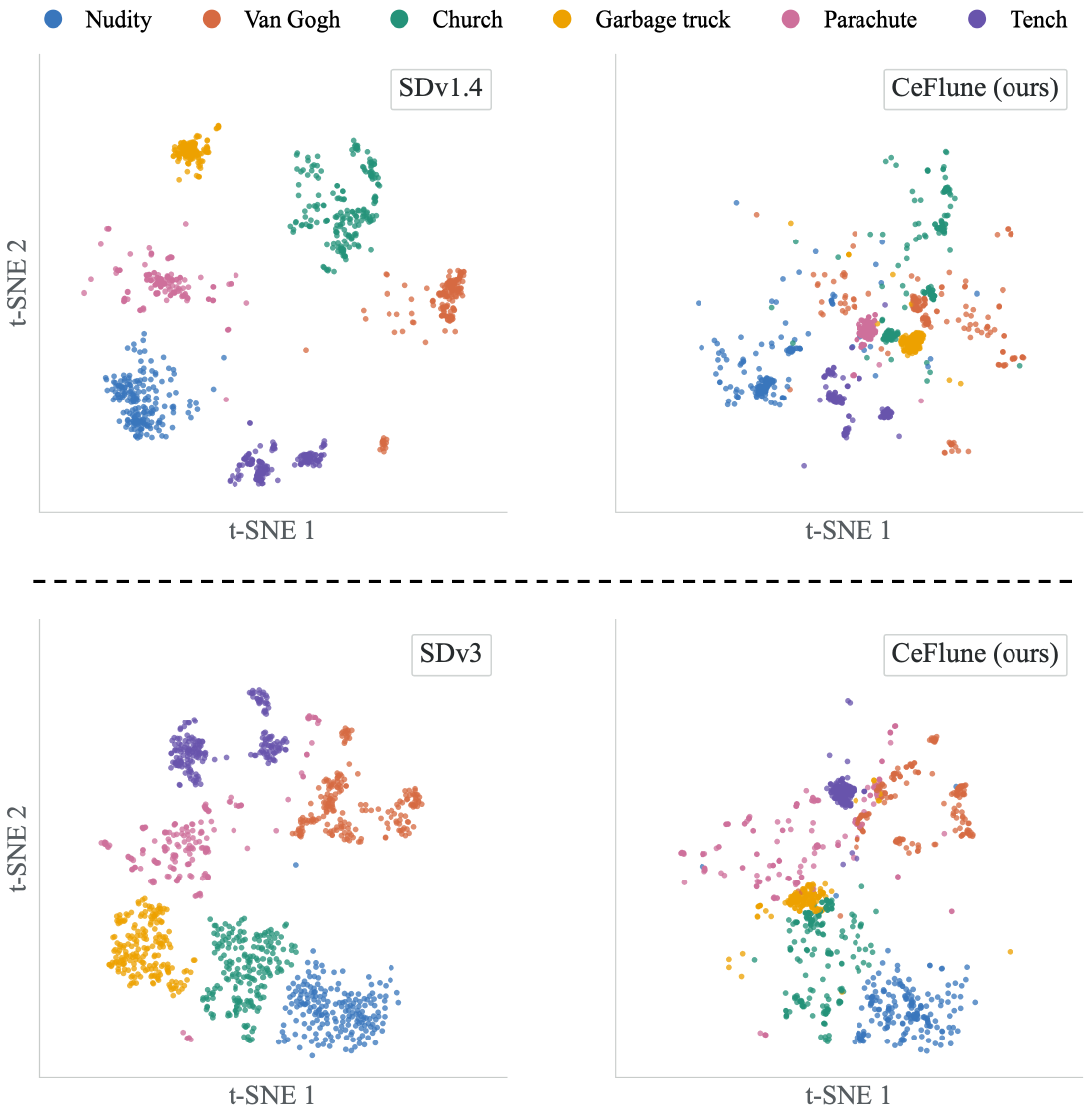}
    \caption{t-SNE visualization of generated images before and after concept erasure on SDv1.4 and SDv3. Within each model architecture, CLIP image features from all evaluated models are projected together.}
    \label{fig:e15_distribution}
\end{figure}

\subsubsection{t-SNE Visualization of Erasure Effects}
We used the evaluation prompts for the six core concepts described in Section 5.1.1 to obtain images from the original models and their counterparts after CeFlune erasure. For each model architecture, we used the CLIP image encoder to encode images from all evaluated models and then jointly projected their features into two dimensions using t-SNE \cite{van2008visualizing}. Fig.~\ref{fig:e15_distribution} shows that the original models’ images form distinct clusters for the six concepts. After CeFlune erasure, some of the corresponding clusters shift from their original locations, and many move toward the center of the projection, suggesting that visual features associated with the concepts are weakened.

\section{Conclusion}
We introduce initial latent variable optimization (IVO), which extends concept reactivation attacks to the initial latent space and exposes a blind spot in erasure robustness evaluation. We also present CeFlune, a concept erasure method that simulates brain functional localization in text-to-image models. CeFlune uses task–rest response comparison to select concept-responsive Unit Neurons and updates their parameters to erase the target concept. On SDv1.4 and SDv3, comparisons with 15 and 11 erasure baselines, respectively, show that CeFlune achieves a favorable balance between concept erasure and general image generation capability. Across 5 evaluated diffusion and flow-matching models, its mean erasure rate across 6 concepts ranges from 74.6\% to 96.8\%. For the 5 attacks evaluated on SDv1.4, CeFlune also has a lower mean attack success rate than all 8 erasure baselines. These results show that CeFlune can erase concepts across different model architectures and make their reactivation more difficult under attacks.

Future work could extend the Unit Neuron definition beyond linear modules and compare functional localization across different computational scales. It is also worth extending the cross-concept overlap analysis to map how concept-related responses are distributed and shared across the model, providing a clearer basis for selecting intervention sites. Beyond choosing where to intervene, the reappearance of Van Gogh style after LoRA fine-tuning suggests that future methods should allow for changes in requirements after concept erasure.

\section{Acknowledgment}
This work was supported by Beijing Natural Science Foundation (JQ24022), the National Natural Science Foundation of China (No. 62192785, No. 62372451, No. 62372082, No. 62625311, U24A20331, 62536002), Young Elite Scientists Sponsorship Program by CAST (2024QNRC001), and Xinjiang Leading Talents Introduction Program (No. XJRC-2025-KJ-YJ-CXPT-182).

\bibliographystyle{IEEEtran} 
\bibliography{acmart} 

\begin{IEEEbiography}[{\includegraphics[width=1.0in,height=1.1in,clip,keepaspectratio]{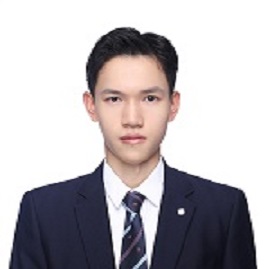}}]{Manyi Li}
received the B.E. degree in artificial intelligence from Beijing Jiaotong University, Beijing, China, in 2025. He is currently pursuing the Ph.D. degree with the School of Advanced Interdisciplinary Sciences, University of Chinese Academy of Sciences, Beijing, China. His research interests include computer vision, generative models, and trustworthy artificial intelligence.
\end{IEEEbiography}

\begin{IEEEbiography}[{\includegraphics[width=0.9in,height=1.05in,clip,keepaspectratio]{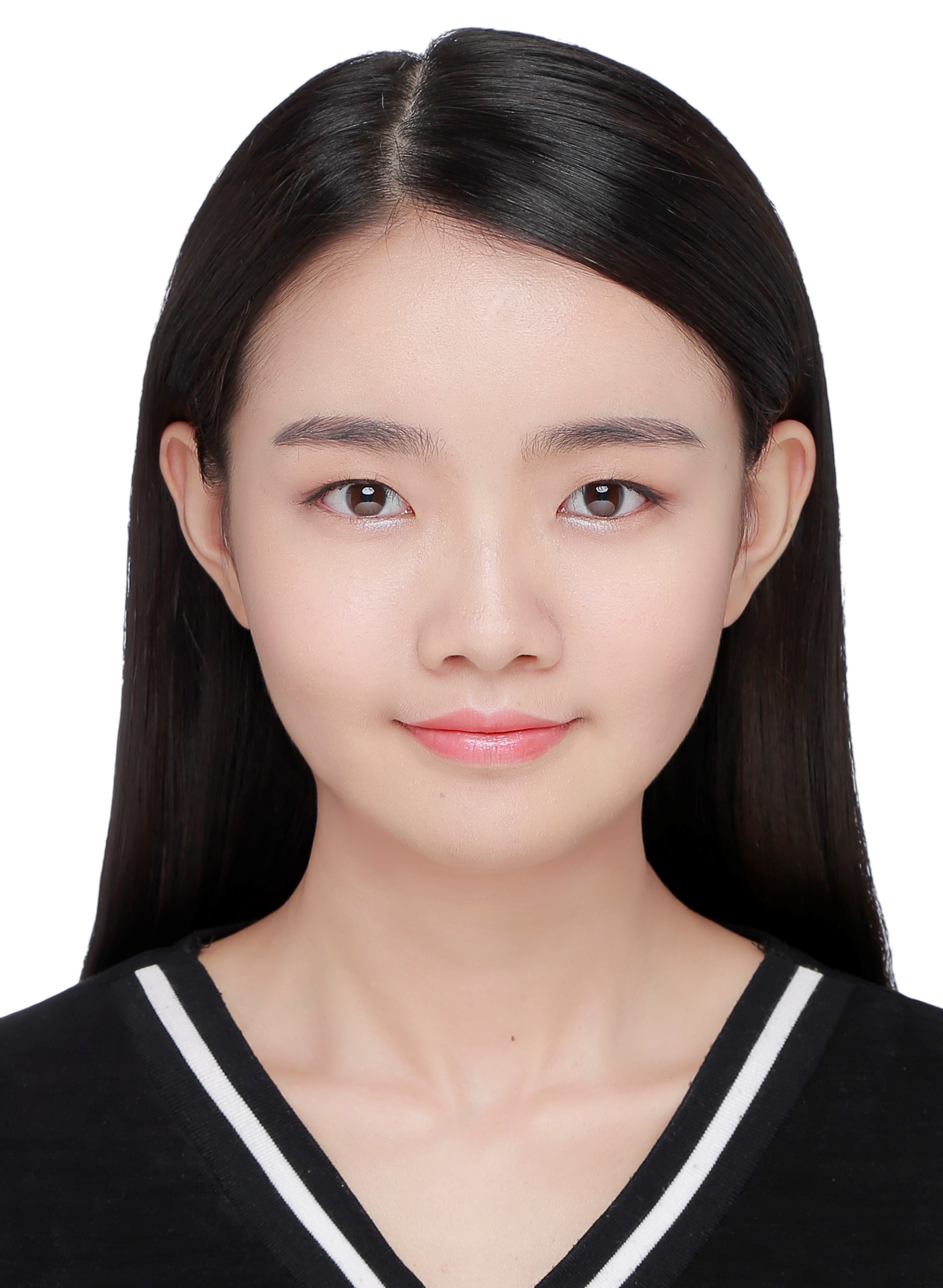}}]{Yufan Liu}
	received the Ph.D. degree from the School of Artificial Intelligence, University of Chinese Academy of Sciences, Beijing, China,  in 2023. She is currently an associate professor with the Institute of Automation, Chinese Academy of Sciences, Beijing, China. Her research interests include computer vision, model compression and multi-modal content understanding. 
\end{IEEEbiography}

\begin{IEEEbiography}[{\includegraphics[width=0.9in,height=1.1in,clip,keepaspectratio]{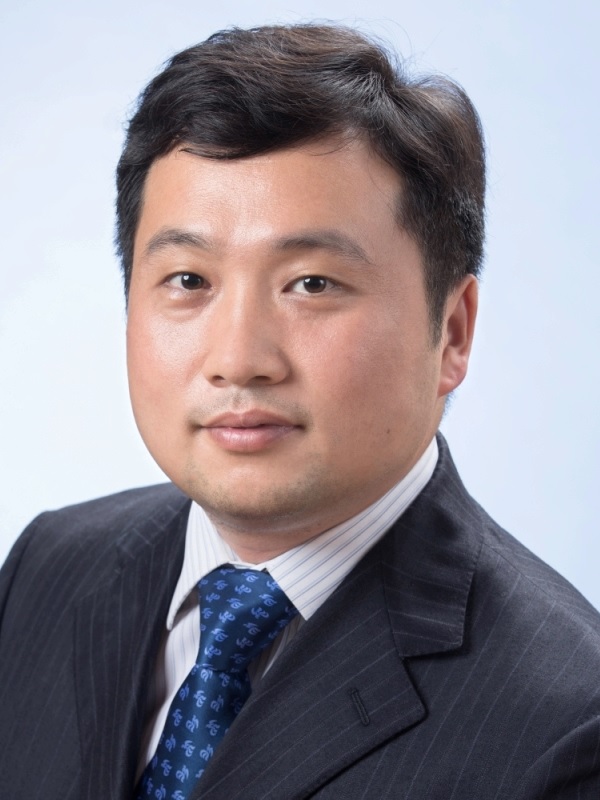}}]{Bing Li}
	received the Ph.D. degree from the Department of Computer Science and Engineering, Beijing Jiaotong University, Beijing, China, in 2009. He is currently a Professor with the Institute of Automation, Chinese Academy of Sciences, Beijing. His current research interests include video understanding, color constancy, visual saliency, multi-instance learning, and Web content security.	
\end{IEEEbiography}

\begin{IEEEbiography}[{\includegraphics[width=0.9in,height=1.1in,clip,keepaspectratio]{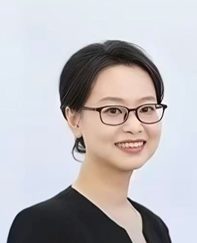}}]{Xia Wu}
received the Ph.D. degree in basic psychology from the State Key Laboratory of Cognitive Neuroscience and Learning, Beijing Normal University, Beijing, China, in 2008. She is currently a Distinguished Professor with the School of Computer Science and Technology, Beijing Institute of Technology, Beijing, China. She has authored more than 100 high-impact journal articles in the past five years, including publications in top-tier venues. Her research interests include artificial intelligence, brain-inspired algorithms, brain-signal analysis, and brain–computer interfaces.
\end{IEEEbiography}

\begin{IEEEbiography}[{\includegraphics[width=0.9in,height=1.1in,clip,keepaspectratio]{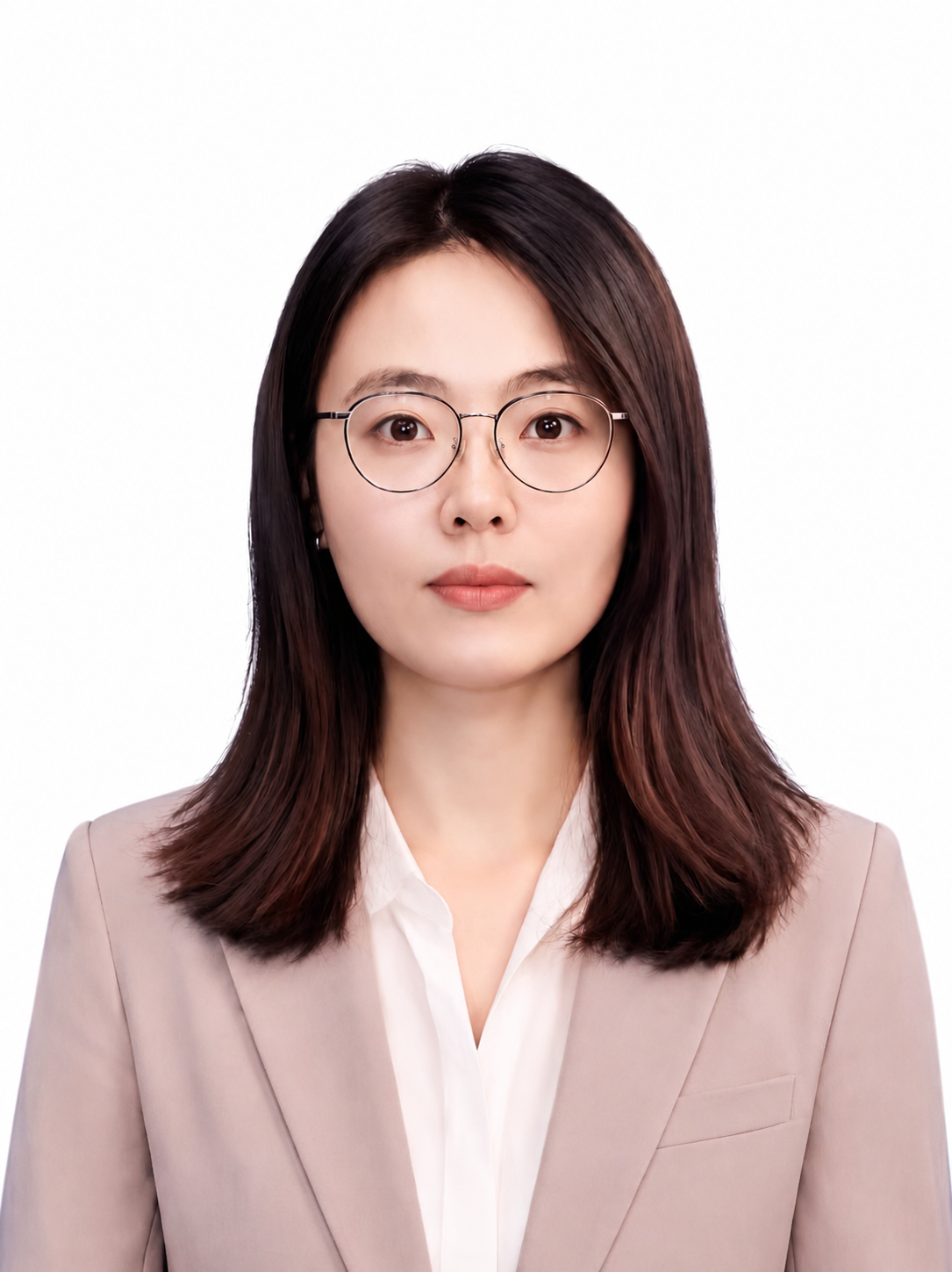}}]{Ziyu Li} received the Ph.D. degree in computer science from Beijing Normal University, Beijing, China. She is currently an Associate Professor with the School of Computer Science and Technology, Beijing Institute of Technology, Beijing, China. Her research interests include brain-signal analysis, brain-computer interfaces, brain-inspired intelligence, and artificial intelligence. \end{IEEEbiography}

\begin{IEEEbiography}[{\includegraphics[width=0.9in,height=1.1in,clip,keepaspectratio]{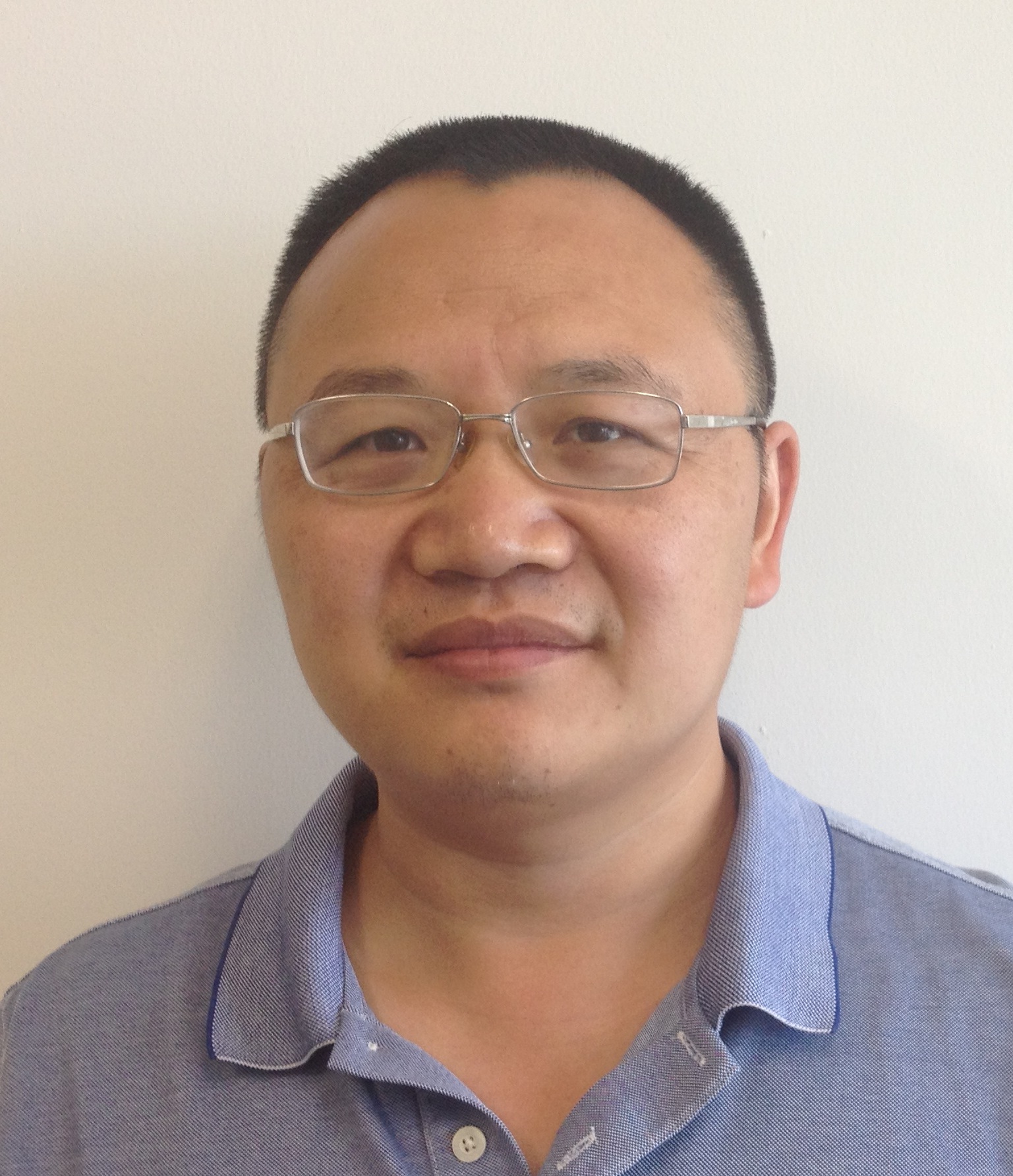}}]{Weiming Hu}
	received the Ph.D. degree from the Department of Computer Science and Engineering, Zhejiang University, Zhejiang, China, in 1998. From 1998 to 2000, he was a postdoctoral research fellow with the Institute of Computer Science and Technology, Peking University, Beijing. He is currently a professor with the Institute of Automation, Chinese Academy of Sciences(CASIA), Beijing. His research interests are visual motion analysis, recognition of web objectionable information, and network intrusion detection.	
\end{IEEEbiography}

\end{document}